\documentclass[journal=gmj]{CUP-JNL-DTM}%

\usepackage{graphicx}
\usepackage{multicol,multirow}
\usepackage{amsmath,amssymb,amsfonts}
\usepackage{mathrsfs}
\usepackage{amsthm}
\usepackage{rotating}
\usepackage{appendix}
\usepackage{ifpdf}
\usepackage[T1]{fontenc}
\usepackage{newtxtext}
\usepackage{newtxmath}
\usepackage{textcomp}
\usepackage{amsthm}
\usepackage{xcolor}
\usepackage{lipsum}
\usepackage[colorlinks,allcolors=blue]{hyperref}

\theoremstyle{definition}

\numberwithin{equation}{section}

\jname{Research Article}
\jyear{2026}

\begin{document}

\pagenumbering{gobble}
\pagestyle{empty}

\begin{Frontmatter}

\title[Article Title]{Atomic Units of X: The Compression Layer of Intelligence}

\author[1]{Sachin Dev Duggal}
\author[1]{Pradyumna Swarnalatha Ramanna}
\author[1]{Alexandros Vassiliades}

\authormark{Sachin Dev Duggal \textit{et al}.}

\address[1]{\orgname{SeKondBrain AI Labs}, \orgaddress{\city{London}, \country{United Kingdom}}}

\authormark{Sachin Dev Duggal et al.}


\abstract{This paper proposes a theoretical and empirical framework for understanding intelligence as a process of atomic compression and compositional reuse. It argues that scalable cognitive, biological, computational, and organisational systems reduce complexity by organising information into reusable units that can be recombined into higher-order structures. The central contribution is the Compression Calculus, a formal framework for comparing surface evidence with atomic representations and for describing how abstraction can compound across layers. The framework is evaluated on large, multi-source software corpora under an open-world concept model, showing substantial evidence-to-concept consolidation in two large projects, while a smaller third project remains below the target threshold. The analysis further identifies important boundary conditions: consolidation depends on corpus scale, evidence-unit definition, and concept identity, and the observed reduction is driven primarily by within-corpus recurrence rather than cross-source recurrence. The paper also develops a representational account of the meaning gap in contemporary generative systems, describing latent, context-dependent conceptual approximations as soft atoms that lack the persistent identity and compositional constraints of stable atomic units. This motivates an architectural view in which large language models function as dynamic fusion engines that navigate and compose persistent conceptual structures rather than serving as the sole repository of those structures.
}

\end{Frontmatter}

\section{Introduction}
\label{sec:intro}

Contemporary artificial intelligence systems have achieved remarkable performance across language understanding, reasoning, retrieval, and decision-support tasks \cite{lee2009advances,brown2020language,lewis2020retrieval}. However, many current architectures still represent knowledge either at very fine-grained levels, such as tokens, or at relatively coarse levels, such as documents, files, and retrieved passages. Although these units are computationally useful, they do not necessarily correspond to the persistent conceptual structures through which knowledge is organised, reused, and combined. This creates a representational gap: large models may process vast amounts of information, while the concepts through which that information is connected, compressed, transferred, and recombined remain implicit, context-dependent, or difficult to inspect.

This paper addresses this gap by proposing atomic units as an intermediate representational layer for intelligence. Atomic units are reusable and composable primitives that organise complex surface evidence around persistent conceptual structures. The underlying intuition is that intelligence often operates less through repeated processing of raw information than through the recognition and reuse of stable building blocks. Expert chess players perceive board configurations as familiar chunks \cite{miller1956magical,chase1973perception}; physicians organise symptoms and test results into diagnostic patterns; programmers work with functions, modules, interfaces, and design patterns; and mathematical notation compresses families of operations into compact symbolic forms. Across these domains, the identification of appropriate primitives is central to efficient reasoning, transfer, and adaptation, consistent with prior work on chunking, modularity, and compression-based accounts of intelligence \cite{simon1962architecture,solomonoff1964formal,kolmogorov1965three}.

Existing approaches nevertheless provide only a limited account of the intermediate layer between low-level elements and high-level artefacts. Tokens are often too fine-grained to function as stable conceptual units, while documents and fixed-size passages may contain several concepts or distribute one concept across multiple sources. Contemporary generative systems may develop latent representations that approximate conceptual structure, but such representations are generally reconstructed within context and need not possess the persistent identity, compositional stability, or explicit compositional constraints associated with reusable atomic units. This paper refers to these context-dependent approximations as \emph{soft atoms} and uses the distinction to formulate the meaning gap as a representational problem rather than as a claim about the general capabilities of language models.

The paper therefore examines how atomic representations can be formally described, how distributed evidence can be consolidated around persistent concepts, how atomic quality and granularity affect representation, and how stable units can support higher-order abstraction and composition. The Compression Calculus provides the formal framework for this analysis by relating surface evidence to atomic representations under explicitly defined measurement conventions. The paper further introduces the Compounding Cascade thesis, according to which stable abstractions can serve as the substrate for subsequent abstraction layers, allowing representational gains and quality effects to propagate across a hierarchy.

The theoretical framework is complemented by an empirical evaluation of software-engineering corpora constructed from repository evidence and public Apache mailing-list archives. Repository data from SEOSS-33 are combined with communication evidence for HBase, Hadoop, and Maven using a common acquisition, preprocessing, linking, and measurement procedure. The evaluation employs an open-world concept model in which evidence that cannot be linked to an existing issue identifier is retained and may introduce additional thread-level concepts. This permits both the evidence set and the concept set to expand when additional sources are incorporated, rather than restricting the analysis to material that already fits a predefined identifier system.

The empirical analysis is intended to measure evidence-to-concept consolidation rather than to establish that the operational reference concepts constitute fully formed atomic units. The results also expose important boundary conditions. Consolidation varies with corpus scale, evidence-unit definition, and the rule used to establish concept identity. At deduplicated paragraph granularity, the target consolidation level is reached in the two larger projects, HBase and Hadoop, but not in the smaller Maven corpus, and it is not reached at message granularity in any of the three projects. The measurements further indicate that the principal source of the observed consolidation is within-corpus recurrence and the selected unit boundary rather than recurrence introduced specifically by combining repository and mailing-list sources. The multi-source contribution of the evaluation is therefore methodological: it demonstrates how heterogeneous evidence can be retained and organised under a common open-world concept model without attributing the resulting consolidation to cross-source recurrence alone.

\subsection{Motivation}
\label{sub:motivation}

The motivation for this work is the need for knowledge architectures that are not only larger, but also more structured, inspectable, and capable of evolving as new evidence becomes available. Scaling model size or context length does not by itself resolve how knowledge should be organised, updated, related across sources, or reused over time. Nor does increasing the quantity of retrievable material determine the representational unit through which repeated evidence should be identified as concerning the same underlying concept.

Systems based primarily on token-level prediction or document-level retrieval may therefore lack an explicit intermediate conceptual layer at which evidence concerning the same subject can be consolidated. A software issue, for example, may be represented across an issue description, several comments, commits, and mailing-list discussions. Treating these artefacts as unrelated retrieval units preserves their surface form but leaves their shared conceptual identity implicit. Conversely, representing the entire collection as a single coarse unit would remove distinctions needed for provenance, temporal analysis, and reuse.

The central representational problem is consequently one of granularity. Units that are too fine require higher-level meaning to be reconstructed repeatedly from surface fragments, whereas units that are too coarse reduce precision and compositional flexibility. A useful intermediate layer must preserve sufficient structure for reasoning while remaining stable enough to be referenced, reused, and combined across contexts.

This paper argues that persistent atomic concepts can provide such an organisational layer. Original evidence remains available for provenance, temporal context, and technical detail, while atomic units provide reference structures through which related material can be connected and reused. This distinction is particularly relevant to retrieval-augmented systems, knowledge graphs, semantic memory, neuro-symbolic architectures, and agentic systems that must maintain and refine knowledge beyond a single interaction.

The same distinction also motivates a separation between knowledge representation and knowledge navigation. Rather than requiring a generative model to act simultaneously as the repository of domain structure and the mechanism that navigates it, the framework developed here suggests an architecture in which persistent atomic structures provide the representational substrate and large language models operate as dynamic fusion engines that select, traverse, and compose those structures for a particular task. The implications of this separation are developed later in the paper.

\subsection{Contributions}
\label{sub:contribution}

The paper makes the following contributions:

\begin{itemize}

\item It introduces atomic units as reusable and composable primitives for organising evidence across cognitive, computational, biological, and organisational systems.

\item It proposes the Compression Calculus, a formal framework for analysing the relationship between surface evidence and atomic representation.

\item It defines measures for reasoning about evidence-to-concept consolidation, atomic compression, domain-level compression, compositional gain, and compression across abstraction layers.

\item It introduces the Compounding Cascade thesis, according to which stable units can provide the basis for progressively higher levels of abstraction and through which representational gains and quality effects can compound across layers.

\item It develops an open-world concept model in which evidence that cannot be linked to an existing identifier is retained and may introduce additional concepts.

\item It evaluates evidence-to-concept consolidation across three software projects using SEOSS-33 repository data and public Apache mailing-list archives for HBase, Hadoop, and Maven, applying a common open-world, cross-source measurement procedure while explicitly analysing the effects of corpus scale, evidence-unit definition, and concept identity.

\item It develops the notion of \emph{soft atoms} to characterise context-dependent conceptual approximations in generative systems and uses compositional stability, persistent identity, and compositional constraints to distinguish such approximations from stable atomic units.

\item It applies the Compression Calculus analytically to worked cross-domain cascades, illustrating how successive abstraction layers may compound under stated assumptions while distinguishing these estimates from empirical measurements.

\item It formulates a set of testable predictions concerning explicit atomic libraries, entropy-bounded compression, multiplicative quality effects, library growth, concept-organised retrieval, and the scale of expert conceptual libraries.

\item It develops implications for knowledge representation and AI architecture, particularly for retrieval-augmented generation, semantic memory, neuro-symbolic systems, and the role of large language models as dynamic fusion engines operating over persistent conceptual structures.

\end{itemize}

\subsection{Scope}
\label{sub:limitations}

The empirical evaluation focuses on software-engineering corpora in which repository evidence and developer communication can be connected through persistent work-item identifiers and reply-thread structure. Three Apache projects are examined: HBase, Hadoop, and Maven. The resulting reference concepts are operationally grounded primarily in software issues and discussion threads. The evaluation therefore measures evidence consolidation within this setting rather than claiming that the same concept definitions, unit boundaries, or consolidation ratios apply directly to every domain.

The operational concepts used in the evaluation should also be distinguished from the stronger theoretical definition of an atomic unit developed in the paper. JIRA issue identifiers provide persistent identity, while thread-level concepts provide a weaker form of continuity for otherwise unlinked discussions. The evaluation does not establish explicit compositional constraints for these concepts, nor does it estimate their atomic quality relative to an optimum. Its empirical contribution is consequently a measurement of consolidation around persistent reference structures rather than a demonstration that the resulting concepts satisfy every property of a fully developed atomic library.

The evaluation further identifies boundary conditions on consolidation. The target consolidation level is reached at deduplicated paragraph granularity in the two larger software corpora but not in the smaller Maven corpus, and substantially weaker consolidation is observed in contrasting domains examined later in the paper. Corpus scale, recurrence structure, unit definition, and concept-identity rules are therefore part of the empirical result rather than incidental preprocessing choices. In particular, the multi-source design should not be interpreted as evidence that cross-source recurrence itself produces the observed reduction; the analysis distinguishes the contribution of an additional evidence source from recurrence already present within the underlying corpus.

The identification of atomic units more generally depends on the task and on the structure of the available evidence. Units that are too narrow may fragment meaningful structures and impose excessive composition costs, while units that are too broad may represent individual cases compactly but provide limited opportunities for reuse and recombination. The framework therefore treats atomic granularity as a domain-dependent design decision rather than as a universal fixed level.

Finally, compression should not be interpreted as the deletion or semantic replacement of original evidence. Documents, messages, issues, commits, comments, and other source artefacts remain necessary for provenance, temporal context, and domain-specific detail. Atomic units provide an organisational layer through which this evidence can be connected and reused rather than a substitute for the underlying sources. Likewise, the empirical evaluation measures consolidation and does not directly establish semantic fidelity, retrieval quality, automatic atom discovery, or compositional gain.

The remainder of the paper is organised as follows. Section~\ref{sec:related} reviews related work on cognitive chunking and expertise, information-theoretic views of intelligence, modularity and evolvability, symbolic and compositional representation, retrieval-augmented generation, neuro-symbolic AI, and library learning. Section~\ref{sec:background} develops the theoretical background of the framework, including atomic units, compression, symbolic notation, cognitive chunking, compositionality, and the distinction between stable atomic units and soft atoms in contemporary generative systems. Section~\ref{sec:methodology} introduces the Compression Calculus, including surface and atomic representations, evidence-to-concept consolidation, compositional compression, information-theoretic bounds, the Compounding Cascade, atomic granularity and quality, the application protocol, and analytical cross-domain cascade estimates. Section~\ref{sec:evaluation} presents the three-project software evaluation, its boundary conditions, comparisons with independently published empirical anchors, and the reproducibility materials. Section~\ref{sec:discussion} examines the implications for knowledge representation, retrieval and semantic memory, large language models as dynamic fusion engines, evolving atomic libraries, and the testable predictions derived from the framework. Finally, Section~\ref{sec:conclusion} concludes the paper.
\section{Related Work}
\label{sec:related}

This work builds on several lines of prior research that, despite originating in different disciplines, converge on a common theme: intelligent systems often manage complexity by organising recurring patterns into reusable and composable units. Relevant prior work spans cognitive psychology, information theory, modularity and evolvability, symbolic and linguistic representation, retrieval-augmented language models, neuro-symbolic AI, and program synthesis through library learning. These traditions provide complementary accounts of compression, abstraction, reuse, and composition, but generally address them independently. This section reviews these foundations and identifies the representational gap addressed by the framework developed in the remainder of the paper.

\subsection{Chunking, Expertise, and Cognitive Compression}

The cognitive-science literature provides one of the earliest and most direct accounts of compressed units in intelligent behaviour. Miller's classic work on the limits of short-term memory argued that human working memory is constrained not simply by raw information quantity, but by the number of meaningful ``chunks'' that can be held and manipulated at once \citep{miller1956magical}. Subsequent studies of expertise showed that experts do not merely store more facts than novices. Instead, domain-specific experience becomes organised into higher-level patterns that can be recognised and manipulated as coherent units.

Chess expertise has been particularly influential in this respect. Chase and Simon showed that skilled chess players recall meaningful board positions substantially better than novices, while much of this advantage disappears for randomly arranged positions \citep{chase1973perception}. Their results supported the view that expert performance depends on structured pattern recognition rather than superior general memory. Later work by Gobet and Simon developed this account through chunking and template theory. Estimates discussed in this literature place the long-term memory repertoire of a chess master at roughly 50,000 or more familiar chunks \citep{gobet1996templates}. Experimental work further showed that expert chunks may be considerably larger than suggested by the original Chase--Simon experiments: for Masters, Gobet and Simon reported a mean of the median largest chunk size of 16.8 pieces in a recall task \citep{gobet1998expert}. These quantities should be interpreted as domain-specific results rather than as universal constants, but they illustrate the scale and structure that learned representational libraries can acquire with expertise.

Similar principles appear in other expert domains. Physicians recognise diagnostic configurations rather than treating symptoms and test results as independent observations; programmers employ code idioms, functions, interfaces, and design patterns; radiologists recognise diagnostically meaningful visual configurations; and musicians reason through motifs, phrases, harmonic structures, and progressions. Cognitive architectures such as ACT-R and SOAR likewise employ chunk-like structures as important representational components \citep{anderson1996act,laird2012soar}.

Further evidence comes from expert--novice studies outside chess. In physics problem solving, Chi, Feltovich, and Glaser showed that novices tend to categorise problems according to surface features, whereas experts group them according to underlying physical principles \citep{chi1981categorization}. Studies of medical reasoning similarly suggest that experienced clinicians organise patient information through structured causal and diagnostic models rather than treating observations as isolated facts \citep{patel1986knowledge}. Research on radiological expertise also indicates that expert performance depends on recognising and organising complex visual patterns into diagnostically meaningful structures \citep{lesgold1988expertise}. Together, these findings support the broader proposition that expertise involves the acquisition of domain-specific representational units that reduce the effective complexity of reasoning.

\subsection{Compression, Prediction, and Information-Theoretic Accounts of Intelligence}

A second foundation comes from information theory and algorithmic accounts of prediction and representation. Solomonoff induction and Kolmogorov complexity connect prediction with the search for compact descriptions of observed data \citep{solomonoff1964formal,kolmogorov1965three}. Under this perspective, recurring structure permits a phenomenon to be represented more economically than its unstructured surface form. Compression is therefore related not only to storage efficiency but also to the identification of regularities that support prediction and generalisation.

Work connecting compression and intelligence develops this relationship further. Hutter considers compression as closely related to general predictive capability, while Schmidhuber's theory of compression progress links improvements in representation to curiosity and the discovery of previously unrecognised regularities \citep{hutter2006prize,schmidhuber2009compression}. Friston's free-energy principle provides a related biological perspective by modelling perception and action in terms of predictive processes that minimise surprise \citep{friston2010free}. Although these approaches differ substantially in formalism and scope, they share the idea that useful internal representations exploit regularity rather than repeatedly processing observations as independent events.

The Minimum Description Length tradition makes this principle explicit in statistical model selection. MDL seeks representations that balance the complexity of the model against the cost of describing the observations through that model \citep{rissanen1978modeling}. This is particularly relevant to the problem of representational granularity: a representation can become ineffective if the vocabulary required to encode it is itself excessively costly. The information bottleneck method develops a related principle by seeking compressed representations that preserve information relevant to a target variable \citep{tishby2000information}. These approaches therefore connect compression to the preservation of task-relevant structure rather than to reduction in size alone.

Recent machine-learning research has strengthened the connection between predictive modelling and compression. Del{\'e}tang et al.~\citep{deletang2024language} demonstrate that predictive language models can be interpreted as general-purpose compressors and show that compression performance can provide insight into properties such as scaling and representation. This work reinforces the view that compression is closely connected to learned structure. However, information-theoretic and predictive accounts do not by themselves determine the representational granularity at which persistent knowledge should be stored, nor do they specify how reusable compressed units should compose into higher-order structures.

\subsection{Modularity, Compositionality, and Evolvability}

The literature on complex systems provides a third foundation. Simon's account of nearly decomposable systems argues that complex systems become more tractable and adaptable when organised into relatively stable intermediate structures \citep{simon1962architecture}. His watchmaker parable illustrates how systems assembled from reusable subcomponents can be more resilient to interruption and change than systems constructed monolithically. The broader principle has influenced theories of hierarchy, modularity, and system design across cognitive science, biology, and engineering.

Evolutionary biology provides a parallel argument. Wagner and Altenberg's work on modularity and evolvability describes how modular organisation can facilitate adaptation by allowing components to vary and recombine without requiring simultaneous changes to an entire system \citep{wagner1996complex}. Biological structures such as protein domains provide examples of functional units that recur within different larger structures. Similar organisational principles appear in engineered systems, where functions, interfaces, components, and services permit complex systems to be constructed from relatively stable reusable elements.

Software engineering has developed these principles into explicit design practices. The Gang of Four design patterns provide a shared vocabulary of reusable software abstractions \citep{gamma1994design}. Atomic design proposes a hierarchy from atoms to molecules, organisms, templates, and pages in user-interface construction \citep{frost2016atomic}. Domain-Driven Design similarly organises software around recurring representational structures such as entities, value objects, aggregates, domain events, repositories, and bounded contexts \citep{evans2003domain}. Workflow research identifies recurring control-flow, data, and resource patterns that can be composed into larger processes \citep{van2003workflow}. Across these traditions, modularity provides mechanisms for reuse, maintainability, local modification, and scalable composition.

Recent machine-learning research has also revived modularity as an architectural principle. Surveys of modular deep learning describe modularity across data, task, and model dimensions and identify potential advantages including interpretability, scalability, recombination, and reuse \citep{sun2023modularity}. Related work treats neural systems as collections of computational modules in which routing mechanisms determine which components should be activated for a particular input or task \citep{pfeiffer2023modular}. This extends classical modularity into learned systems: modules may be acquired, selected, routed, specialised, and recombined rather than being entirely hand-designed.

Modularity alone, however, does not specify what should constitute the representational unit of knowledge. A computational module may encapsulate a transformation or capability without corresponding to a persistent domain concept. The literature therefore establishes the advantages of intermediate structure and reuse while leaving open the question of how the units themselves should be identified and measured.

\subsection{Language, Symbolic Systems, and Compositional Representation}

Language and symbolic systems provide another important body of related work. Formal notation systems, particularly in mathematics, demonstrate how compact symbols can represent procedures and relationships that would otherwise require substantially longer surface descriptions. The progression from natural-language arithmetic to symbolic notation, and from arithmetic to algebra and calculus, illustrates how new representational units can simultaneously compress prior structures and make new forms of manipulation possible.

Linguistics provides related examples of structured representation. Speech act theory categorises utterances according to functions such as assertions, directives, commissives, expressives, and declarations \citep{austin1962things,searle1969speech}. Rhetorical Structure Theory decomposes discourse into elementary discourse units connected by relations such as elaboration, contrast, cause, and condition \citep{mann1988rhetorical}. Technical-documentation frameworks such as DITA organise information into reusable topics and maps \citep{priestley2001dita}. Chomsky's Minimalist Program proposes Merge as a minimal operation from which increasingly complex syntactic structures can be constructed \citep{chomsky1995minimalist}. Wierzbicka's theory of semantic primes similarly investigates whether complex meanings can be expressed through combinations of a more restricted set of primitive meanings \citep{wierzbicka1996semantics}. Although these approaches differ substantially in purpose, they illustrate the explanatory power of recurring units governed by explicit compositional relations.

Recent AI research has reframed compositionality as an empirical generalisation problem. Rather than asking only whether complex expressions can theoretically be decomposed into simpler components, this literature examines whether learned models can systematically recombine familiar elements in configurations not observed during training. Surveys of compositional learning review benchmarks, model classes, and evaluation procedures designed to test this capability \citep{sinha2024survey,lin2023survey}. Grounded-language and knowledge-graph experiments likewise show that language models and graph-augmented systems can struggle with sequences of unseen lengths and novel combinations of familiar components \citep{wold2024compositional}. The resulting distinction is important for knowledge representation: the presence of distributed learned features does not by itself guarantee that their meaning remains stable when those features are recombined.

\subsection{RAG, LLMs, and Knowledge Representation Granularity}

Contemporary language-model architectures make the question of representational granularity particularly explicit. Large language models operate over token sequences, whereas retrieval-augmented generation systems typically retrieve larger textual units such as passages, chunks, documents, or files. Both levels are useful computationally, but neither necessarily corresponds to the persistent conceptual structure of the underlying domain. Tokens are usually below the level of complete concepts, while fixed-size passages and documents may contain multiple concepts or distribute a single concept across several sources.

This problem has motivated growing research on retrieval-augmented generation, graph-enhanced retrieval, persistent memory, and structured knowledge integration. Standard RAG improves access to external evidence by retrieving passages relevant to an input query \citep{lewis2020retrieval}. GraphRAG and related approaches introduce graph structure over collections of textual evidence to support query-focused summarisation and corpus-level reasoning \citep{edge2024local}. Tool-using systems connect language models to external operations or structured resources, allowing some information and functionality to remain outside the model's parameters \citep{schick2023toolformer,patil2023gorilla}. These architectures demonstrate the value of separating generation from at least part of the knowledge-access process.

Recent RAG variants make the granularity problem more explicit. RAPTOR recursively embeds, clusters, and summarises text to construct a hierarchy that supports retrieval at different levels of abstraction \citep{sarthi2024raptor}. Self-RAG addresses a complementary problem by training models to determine when retrieval is appropriate, incorporate retrieved evidence, and critique generated responses \citep{asai2024selfrag}. Such approaches show that retrieval performance depends not only on whether external information is available, but also on how that information is segmented, abstracted, selected, and incorporated into reasoning.

Critiques of large language models raise a related representational question. Bender et al.~\citep{bender2021dangers} argue that statistical language modelling over linguistic form does not itself provide grounded meaning. Scaling-law studies show that model performance can improve predictably with increasing data, parameters, and computation \citep{kaplan2020scaling}, but scaling alone does not determine whether the operative representations correspond to persistent, independently addressable, and compositionally constrained concepts. Conversely, successful behaviour by contemporary models provides evidence that distributed neural representations can encode substantial structural information. The unresolved issue is therefore not simply whether useful representations are learned, but what properties such representations possess and whether they can provide stable units for persistent reuse and composition. This representational question is developed further in Section~\ref{sec:background}.

\subsection{Neuro-Symbolic AI and Library Learning}

Neuro-symbolic AI provides a final closely related foundation. Neural approaches provide powerful mechanisms for learning representations from data, while symbolic systems offer explicit structure, constraints, and rule-governed composition. Neuro-symbolic research attempts to integrate these capabilities, particularly where systematic reasoning, interpretability, and compositional generalisation are required \citep{garcez2023neurosymbolic}.

The compositionality problem is visible in benchmarks such as SCAN, where sequence-to-sequence models have historically struggled to apply familiar components systematically in combinations outside their training distribution \citep{lake2018generalization}. Neural-Symbolic Stack Machines address this limitation by combining neural prediction with an explicitly executable symbolic stack mechanism, enabling recursive and systematic composition \citep{chen2020neural}. Logic Tensor Networks integrate neural learning with differentiable logical constraints \citep{badreddine2022logic}, while Neuro-Vector-Symbolic Architectures combine learned perception with structured vector-symbolic operations for abstract reasoning tasks \citep{hersche2023neuro}. These approaches provide evidence that explicit compositional structure can complement learned distributed representations.

Program synthesis and library learning address a closely related question from a different direction: whether reusable abstractions can themselves be discovered rather than supplied in advance. DreamCoder learns libraries of increasingly abstract program components through a wake--sleep procedure in which program synthesis and library construction reinforce one another \citep{ellis2021dreamcoder}. This is particularly relevant to the present work because abstraction discovery becomes part of the learning process. Reusable units are not assumed to be fixed permanently; they can emerge from recurring structures encountered while solving tasks.

More recent work connects this principle directly to language-model-based systems. LILO combines LLM-guided program synthesis with symbolic compression and automatic documentation to construct interpretable libraries of reusable abstractions \citep{grand2024lilo}. Voyager applies a related principle in an embodied-agent setting by maintaining an expanding library of executable skills that can be retrieved, refined, and reused across tasks \citep{wang2023voyager}. These systems demonstrate that persistent libraries can complement generative models by preserving abstractions outside any single inference context.

Library-learning approaches are consequently among the closest precedents for the architecture considered in this paper. Their primary objects, however, are executable programs or skills, and their compression objectives are typically defined over program libraries and synthesis tasks. They do not directly provide a general account of how heterogeneous surface evidence should be consolidated around persistent concepts, how representational and composition costs should be compared across domains, or how compression should be analysed across successive abstraction layers.

Taken together, these literatures provide substantial evidence that intelligence and scalable knowledge systems benefit from compression, chunking, modularity, compositionality, structured retrieval, symbolic constraints, and learned libraries of reusable abstractions. They also expose a common unresolved problem. Existing approaches typically focus on one part of the representational process: discovering patterns, compressing observations, constructing modules, retrieving passages, imposing compositional rules, or learning reusable programs. What remains underdeveloped is a common analytical account linking surface evidence to persistent reusable units, incorporating the cost and quality of their composition, and describing how such representations can support successive abstraction layers. The following sections develop such an account through the concepts of atomic representation, the Compression Calculus, and the Compounding Cascade.

\section{Theoretical Background}
\label{sec:background}

This section introduces the theoretical concepts that support the proposed framework. The aim is not yet to present the Compression Calculus itself, which is introduced in Section~\ref{sec:methodology}, but to establish the background assumptions on which it relies. The section focuses on six ideas: atomic units and abstraction, compression as representation, symbolic notation as an intuitive example of compression, chunking in expert cognition, compositionality across abstraction layers, and the representational distinction between persistent atomic units and the context-dependent approximations developed by contemporary generative systems.

\subsection{Atomic Units and Abstraction}
\label{sub:atomic-units-abstraction}

Complex cognitive and computational systems rarely operate directly on raw, undifferentiated information. Instead, they tend to organise information into intermediate units that can be recognised, reused, and recombined. In this paper, these intermediate units are referred to as \emph{atomic units}: compact representational primitives that capture recurring structures within a domain. Atomic units are not necessarily minimal in a physical or mathematical sense. Rather, they are atomic because they function as meaningful building blocks at the level at which a system reasons, learns, retrieves information, or acts.

The theoretical motivation for atomic units is closely related to the study of modular and nearly decomposable systems. Simon's account of the architecture of complexity argues that complex systems become manageable when they are organised into substructures with stronger internal interactions than external interactions \cite{simon1962architecture}. Such modular organisation allows complex systems to be constructed, maintained, and adapted without requiring every element to be reconsidered at once. In cognitive and knowledge-based systems, atomic units play a similar role: they provide stable intermediate structures between surface-level evidence and higher-level reasoning.

This view also connects to evolutionary accounts of modularity and evolvability. Biological systems frequently reuse structural and functional modules, and evolutionary theory has argued that such modularity facilitates adaptation by allowing useful components to be recombined in new contexts \cite{wagner1996complex}. By analogy, knowledge systems that represent information as reusable primitives may support more efficient transfer, recomposition, and adaptation than systems that store only raw observations or large, weakly structured documents.

\begin{figure}[t]
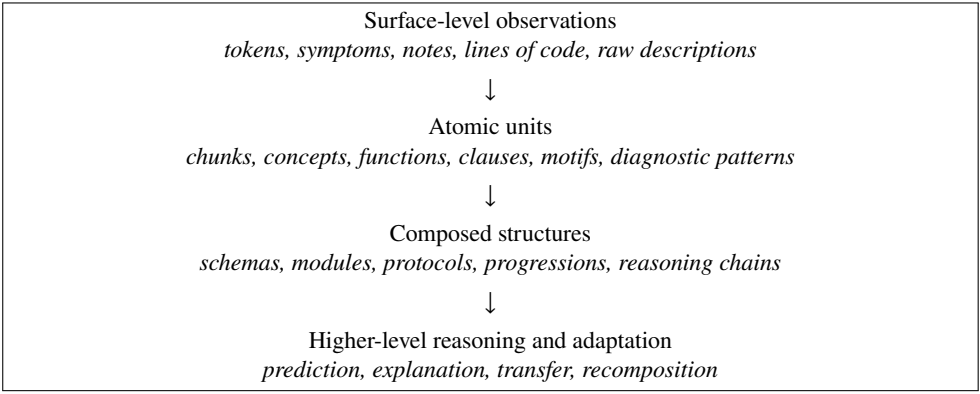

\centering
\fbox{%
\begin{minipage}{0.88\textwidth}
\centering
\small

Surface-level observations\\
\emph{tokens, symptoms, notes, lines of code, raw descriptions}\\[0.4em]

$\downarrow$\\[0.4em]

Atomic units\\
\emph{chunks, concepts, functions, clauses, motifs, diagnostic patterns}\\[0.4em]

$\downarrow$\\[0.4em]

Composed structures\\
\emph{schemas, modules, protocols, progressions, reasoning chains}\\[0.4em]

$\downarrow$\\[0.4em]

Higher-level reasoning and adaptation\\
\emph{prediction, explanation, transfer, recomposition}

\end{minipage}%
}
\caption{Conceptual view of atomic units as intermediate representations between surface-level evidence and higher-level reasoning.}
\label{fig:atomic-abstraction-stack}
\end{figure}

Figure~\ref{fig:atomic-abstraction-stack} summarises this view. Surface-level observations are organised into atomic units. These units can then be composed into larger structures, which in turn support higher-level reasoning, prediction, explanation, transfer, and adaptation.

\subsection{Compression as Representation}
\label{sub:compression-representation}

A second theoretical foundation is the idea that representation and compression are closely related. A representation is useful when it preserves the relevant structure of a phenomenon while reducing unnecessary detail. From this perspective, abstraction can be understood as a form of selective compression: it removes surface variation while retaining the structure needed for prediction, explanation, or action.

At an intuitive level, the representational advantage of an atomic unit can be expressed as a ratio between a surface-level description and a more compact atomic representation:

\begin{equation}
\mathrm{CR}(x)
=
\frac{\left|r_{\mathrm{surface}}(x)\right|}
     {\left|r_{\mathrm{atomic}}(x)\right|}.
\label{eq:compression-ratio}
\end{equation}

Here, $x$ denotes the phenomenon being represented, $r_{\mathrm{surface}}(x)$ denotes its surface-level representation, and $r_{\mathrm{atomic}}(x)$ denotes its compressed atomic representation. A value greater than one indicates that the atomic representation is more compact than the corresponding surface form. This informal expression anticipates the more precise Atomic Compression Ratio introduced in Section~\ref{sec:methodology}.

The comparison depends on how the representations are defined and measured. Description size may refer to characters, tokens, symbols, records, evidence units, or another domain-appropriate quantity. A compression ratio is therefore meaningful only when the unit definitions and representation boundaries are stated explicitly.

This idea is strongly connected to information-theoretic and algorithmic accounts of intelligence. Solomonoff's theory of inductive inference and Kolmogorov's work on algorithmic complexity link prediction, explanation, and short descriptions of data \cite{solomonoff1964formal,kolmogorov1965three}. A shorter description is not merely more economical; when it captures recurring structure in the data, it may also support generalisation. In this sense, compression is not only a storage mechanism but also a way of expressing regularity in the represented phenomenon.

Related work has extended this intuition to learning and cognition. Schmidhuber argues that progress in compression can explain aspects of curiosity, novelty, and creativity because systems benefit when they discover more compact descriptions of previously irregular data \cite{schmidhuber2009compression}. Friston's free-energy principle similarly frames perception and action as processes that reduce surprise through predictive modelling \cite{friston2010free}. More recently, work on language modelling has connected predictive modelling and compression by interpreting language models as general-purpose compressors \cite{deletang2024language}. These perspectives motivate the claim that intelligent systems benefit from representations that organise recurring surface phenomena into reusable structures. They do not, however, imply that compression ratio alone constitutes a direct measure of understanding; representational fidelity, granularity, and compositional usefulness remain separate considerations in the framework developed below.

\subsection{The Notation Example: From Surface Forms to Symbolic Compression}
\label{sub:notation-example}

Mathematical notation provides a simple and intuitive example of how atomic units compress surface-level descriptions. The expression $1+1$ and the phrase ``one plus one'' express the same operation, but the symbolic form is shorter, easier to manipulate, and more readily composable. The symbol $+$ functions as an atomic unit: it compresses the procedural idea of addition into a compact operator that can be reused across many contexts.

The importance of notation is not limited to saving space. Symbolic forms also make composition easier. For example, the expression
\[
(3 \times 4) + (7 \times 2)
\]
is easier to parse, transform, and combine than its natural-language equivalent. At higher levels, positional notation, arithmetic operators, algebraic variables, exponents, and calculus notation each introduce new representational units that compress the layer below. A digit compresses quantity; an operator compresses a procedure; a variable compresses a family of possible values; and a calculus operator compresses a limiting process.

The following ladder summarises the intuition:

\begin{description}

\item[Level 0: Tally marks.] A number such as 247 requires 247 individual marks.

\item[Level 1: Positional notation.] The numeral $247$ represents the same quantity with three symbols; position becomes part of the representation.

\item[Level 2: Arithmetic operators.] An expression such as $247+389$ compresses a procedural counting operation into a compact symbolic form.

\item[Level 3: Multiplication.] The expression $50\times50$ compresses repeated addition into a single operation.

\item[Level 4: Exponentiation.] The expression $2^{32}$ compresses repeated multiplication into a compact symbolic structure.

\item[Level 5: Algebra.] An expression such as $ax^{2}+bx+c=0$ represents a family of arithmetic problems rather than one specific calculation.

\item[Level 6: Calculus.] An expression such as $\int f(x)\,dx$ compresses a limiting process over many approximating sums.

\item[Level 7: Physical equations.] An equation such as $E=mc^{2}$ compresses a general relationship across a broad class of physical phenomena.

\end{description}

This ladder illustrates a general principle: useful representational units do not only shorten descriptions, but also enable new operations. Positional notation, for instance, does not simply represent numbers more compactly than tally marks or Roman numerals; it also makes arithmetic operations more systematic. Similarly, algebra does not merely abbreviate arithmetic; it makes it possible to reason over families of cases. Calculus does not merely shorten algebraic expressions; it introduces operators for reasoning about change, accumulation, and limits.

The ladder also motivates the idea that compression across abstraction layers is multiplicative rather than merely additive. If one abstraction layer compresses the layer below by a factor $r_{1}$, and a second layer compresses the first by a factor $r_{2}$, then the total compression relative to the surface layer is intuitively proportional to $r_{1}r_{2}$, rather than $r_{1}+r_{2}$. More generally, for a sequence of abstraction layers, the cumulative representational compression can be expressed informally as:

\begin{equation}
\mathrm{CR}_{\mathrm{total}}
\approx
\prod_{i=1}^{n} r_i.
\label{eq:total-compression-ratio}
\end{equation}

This expression is included only as an intuition. The formal treatment of layered compression is introduced later through the Compression Calculus and the Compounding Cascade.

\subsection{Chunking and Expert Cognition}
\label{sub:chunking-expertise}

The cognitive basis for atomic representation is closely related to the theory of chunking. Miller's classic account of short-term memory argues that cognitive capacity is limited not simply by the amount of raw information, but by the number of chunks that can be held and manipulated at once \cite{miller1956magical}. A chunk is a compressed unit of meaning: several lower-level elements are treated as a single higher-level structure. This allows experts to operate with larger effective units than novices.

Chess expertise provides one of the clearest empirical examples. Chase and Simon showed that expert chess players recall meaningful board positions much better than novices, but that this advantage is greatly reduced for random board configurations \cite{chase1973perception}. The expert advantage therefore does not arise from superior raw memory alone; it depends on the ability to recognise meaningful recurring configurations as chunks. Later work by Gobet and Simon developed this idea further through template theory, suggesting that expert memory relies on structured patterns that organise recurring configurations \cite{gobet1996templates}.

The same principle appears across other domains. A physician does not reason over every symptom, test result, and historical detail as isolated data points; rather, clinical expertise involves grouping such observations into diagnostic patterns. A programmer does not reason only in individual lines of code, but in functions, modules, interfaces, and design patterns. A musician does not perceive only isolated notes, but motifs, chords, progressions, and formal structures. In each case, expertise depends in part on learning representational units appropriate to the structure of the domain.

\subsection{Compositionality and Abstraction Layers}
\label{sub:compositionality-abstraction-layers}

Atomic units are useful not only because they compress information, but because they compose. Compositionality means that smaller representational units can be combined into larger structures without requiring the entire surface-level description to be reconstructed each time. This is central to language, mathematics, software engineering, music, and scientific reasoning. A relatively limited vocabulary of meaningful units can generate a much larger space of structures when governed by suitable rules of combination.

Abstraction layers make this compositional process recursive. A lower-level unit can become part of a higher-level unit, and the higher-level unit can itself become an atom for further reasoning. In mathematics, numbers compose into expressions, expressions into equations, equations into models, and models into theories. In software, operations compose into functions, functions into modules, modules into services, and services into systems. In expert cognition, perceptual features compose into chunks, chunks into schemas, and schemas into strategies.

This layered view is important because it explains why representation is not only a matter of storage efficiency. The level at which knowledge is represented determines what operations are available to the system. If knowledge is represented too finely, the system may be required to reconstruct higher-level structure repeatedly from lower-level fragments. If it is represented too coarsely, the representation may lose precision, reusability, or flexibility. The theoretical challenge is therefore to identify the representational grain at which compression, reuse, and valid composition are jointly effective.

This challenge becomes particularly important for contemporary AI systems, which commonly combine token-level processing with document- or passage-level retrieval. The question is not simply whether these systems learn internal abstractions, but whether those abstractions possess the stability and compositional properties required of persistent atomic units. The following subsection develops this distinction before the formal methodology is introduced.

\subsection{The Meaning Gap: Soft Atoms and the Limits of Generative Approximation}
\label{sub:meaning-gap}

The representational granularity discussed above provides a useful perspective on a longstanding question concerning contemporary generative systems. Bender et al.~\cite{bender2021dangers} emphasise the distinction between statistical regularities in linguistic form and grounded meaning, while LeCun's proposed path towards autonomous machine intelligence motivates architectures based on hierarchical predictive representations and world models rather than relying exclusively on autoregressive sequence generation \cite{lecun2022path}. These positions differ in emphasis, but both direct attention towards the level at which information is represented. Within the framework developed here, they can therefore be considered as arguments about representational architecture rather than as categorical judgements about what present language models can or cannot do.

The first issue is granularity. Tokens are computational units defined primarily for efficient sequence processing and vocabulary management. They need not correspond to semantic concepts. A fragment such as ``ion'', for example, occurs within words such as ``opinion'', ``fusion'', and ``orion'', but does not by itself provide a persistent representation of any of those concepts. At the opposite extreme, documents, files, and fixed retrieval passages can contain multiple concepts, while evidence concerning one concept may span several such artefacts. Token-level and document-level representations are therefore useful for computation and storage without necessarily defining the conceptual grain at which persistent knowledge should be organised.

This observation does not imply that transformer models lack internal structure above the token level. Substantial empirical work indicates the opposite. Probing studies have identified syntactic, semantic, and relational information within contextual representations. Tenney et al.~\cite{tenney2019bert}, for example, found that representations within BERT recover information associated with stages of the classical NLP pipeline, including part-of-speech information, parsing, named entities, semantic roles, and coreference. Broader reviews of BERT probing likewise conclude that contextual representations encode substantial linguistic structure, while also emphasising that the interpretation and localisation of that information remain complex \cite{rogers2020primer}. The relevant distinction is therefore not between ``having concepts'' and ``having no concepts'', but between latent, context-sensitive representations and persistent, explicitly addressable units within a knowledge architecture.

This paper uses the term \emph{soft atoms} for conceptual approximations that emerge within distributed generative representations but do not necessarily possess the properties required of stable atomic units. Soft atoms may capture useful semantic regularities and may support successful reasoning or generation. They are ``soft'' because their identity and behaviour are probabilistic and context-dependent rather than being maintained as explicit members of a persistent atomic library.

Three properties distinguish the stronger notion of an atomic unit used in this framework from such approximations.

\begin{enumerate}

\item \textbf{Compositional stability.}
An atomic unit should retain sufficiently stable meaning when it participates in different valid compositions. Its behaviour may depend on context, but the identity of the represented concept and the consequences of permitted combinations should not have to be reconstructed entirely from surface form on every use. Generative representations can exhibit greater sensitivity to wording and contextual configuration, even when different expressions are intended to refer to the same underlying subject. Soft atoms therefore approximate compositional structure without guaranteeing invariant behaviour under semantically equivalent reformulations.

\item \textbf{Persistent identity.}
An atomic unit should remain addressable across documents, interactions, and points in time. New evidence can then be attached to the same reference structure, allowing knowledge to accumulate around a persistent identity. Internal language-model representations are instead generated as part of the computation for a particular context. Even when similar internal patterns recur, the architecture does not ordinarily expose them as stable concept objects to which future evidence can be attached directly. A persistent external library therefore provides a capability that is distinct from latent representation within an individual inference context.

\item \textbf{Compositional constraints.}
An atomic architecture should specify, explicitly or operationally, which combinations of units are permitted or meaningful. Such constraints need not be purely logical; they may be typed relations, schemas, domain rules, temporal restrictions, or learned compatibility conditions. Their purpose is to distinguish a valid composition from one that is merely superficially plausible. A generative model can learn strong statistical preferences over combinations, but statistical plausibility is not equivalent to an explicit composition grammar. From this perspective, some hallucinations can be interpreted as failures of constrained composition: a generated sequence remains locally plausible while selecting or combining conceptual material in a way that is unsupported by the relevant evidence or domain structure.

\end{enumerate}

These properties define an architectural distinction rather than an all-or-nothing theory of understanding. A language model may exhibit sophisticated semantic behaviour while still lacking persistent concept identity outside the inference process. Conversely, an explicit symbolic library may provide persistent identity and strict compositional constraints while lacking the flexibility needed to interpret ambiguous natural language. The distinction therefore concerns which representational properties are supplied by each component of a system.

The Compression Calculus provides a way to express the consequences of this distinction without reducing them to model size or computational cost. Section~4.8 defines the layer-quality factor $q_i$ as the ratio between achieved and optimal layer compression under stated representation and fidelity assumptions. To the extent that unstable identity or weak composition causes a representation to fall short of the available layer structure, the resulting loss can be represented as $q_i < 1$. Equation~(4.36) then shows that quality losses across successive abstraction layers compound multiplicatively. This does not establish a numerical quality factor for present language models, nor is such a quantity estimated in this paper. It instead provides a formal vocabulary for expressing why representational quality at one layer can affect the effectiveness of higher-level compositions.

The same analysis also clarifies the role of persistent external knowledge structures. An explicit atomic library need not replace neural representations. It can instead complement them by providing stable identities, provenance, relations, and compositional constraints that remain available across interactions. Neural models can contribute flexible interpretation, similarity assessment, and context-sensitive selection, while the persistent structure provides continuity and reusable reference objects.

The distinction between soft and persistent atomic representations therefore motivates an architectural separation between representation and navigation. The former concerns the maintenance of a reusable atomic library and its composition rules; the latter concerns identifying which units and compositions are relevant to a particular task. Section~\ref{sub:llmsasdynamic} develops this consequence by considering large language models as dynamic fusion engines operating over persistent conceptual structures.

The theoretical challenge is thus not to replace distributed representation with symbolic representation, nor to claim that one form alone is sufficient for intelligence. It is to determine how stable conceptual identity, learned semantic structure, provenance, and valid composition can coexist within an evolving knowledge architecture. The Compression Calculus introduced in the next section provides the formal framework used to analyse that problem.

x\section{Methodology: The Compression Calculus}
\label{sec:methodology}

This section introduces the Compression Calculus as the analytical methodology of the paper. The framework formalises the relationship between surface evidence, atomic representations, and higher-order abstraction. It defines how representational compression can be measured, how the cost of composition is incorporated, and how compression and representational quality can propagate across successive abstraction layers.

The methodology has two complementary roles. First, it provides a general analytical vocabulary for comparing surface and atomic representations across domains. Second, it defines the evidence-to-concept consolidation measure used in the empirical evaluation in Section~\ref{sec:evaluation}. The latter should be understood as an operational measurement of one structural property relevant to atomic representation rather than as a direct measurement of the complete theoretical construct. Software artefacts and communication fragments are treated as evidence units, while persistent work items and open-world discussion concepts provide the reference layer around which that evidence is organised.

The distinction is important. Atomic compression, evidence consolidation, semantic adequacy, and compositional validity are related but non-equivalent quantities. The empirical evaluation measures the concentration of evidence around persistent references. It does not assume that every operational reference concept already satisfies the stronger properties of compositional stability, persistent identity, and compositional constraints discussed in Section~\ref{sub:meaning-gap}.

\subsection{Framework Overview}
\label{subsec:framework-overview}

The Compression Calculus is based on the distinction between an expanded surface representation and a more compact atomic representation. A surface representation records a phenomenon in the form in which it is directly observed, such as source-code fragments, messages, contract text, clinical observations, musical events, financial records, or natural-language descriptions. An atomic representation organises the same phenomenon through reusable domain-specific units, such as functions, concepts, clauses, diagnostic patterns, motifs, operators, or other persistent reference objects.

The framework does not assume that one universal unit of measurement applies across all domains. Description length may be measured in characters, tokens, records, observations, messages, paragraphs, symbolic operations, or another explicitly defined unit. What is required is that the surface and atomic sides of a comparison are measured consistently and that the unit boundary is stated explicitly.

The analysis proceeds in four stages. First, the surface representation and the candidate atomic representation are defined. Second, their description lengths or unit counts are measured. Third, composition costs and higher-order structures are considered. Fourth, where empirical corpora are available, evidence units are associated with persistent reference concepts and the resulting consolidation structure is measured under an explicitly defined concept model.

These stages correspond to different levels of evidential strength. A direct comparison between paired surface and atomic encodings can support an Atomic Compression Ratio. An evidence-to-concept corpus analysis supports a consolidation ratio. Analytical examples can illustrate how compression may propagate across abstraction layers under stated assumptions. These quantities should not be treated as interchangeable.

\subsection{Surface and Atomic Representations}
\label{subsec:surface-atomic-representations}

Let $D$ denote a domain and let $\phi \in D$ denote a phenomenon in that domain.

\paragraph{Surface representation.}
Let $S_D$ denote the space of surface-level representations in domain $D$. The surface representation of $\phi$ is written as

\begin{equation}
s(\phi) \in S_D.
\label{eq:surface-representation}
\end{equation}

Its description length is denoted by

\begin{equation}
\left|s(\phi)\right|.
\label{eq:surface-length}
\end{equation}

The measurement unit must be defined for each analysis. For example, software may be measured in characters, tokens, lines, messages, or paragraphs; legal text may be measured in words, clauses, or characters; and clinical evidence may be measured in observations or structured fields. The same phenomenon may therefore produce different numerical ratios under different unit definitions. The unit boundary is part of the representation being evaluated and not a neutral preprocessing choice.

\paragraph{Atomic representation.}
Let $A_D$ denote a finite library of atomic units for domain $D$:

\begin{equation}
A_D = \{a_1,a_2,\ldots,a_k\}.
\label{eq:atomic-library}
\end{equation}

An atomic representation of $\phi$ is a composition over elements of $A_D$:

\begin{equation}
\alpha(\phi)
=
C\!\left(a_{i_1},a_{i_2},\ldots,a_{i_m}\right),
\label{eq:atomic-representation}
\end{equation}

where $C$ is a composition operator or grammar rule drawn from a domain-specific grammar $G_D$.

The total description length of the atomic representation is

\begin{equation}
\left|\alpha(\phi)\right|
=
|C|
+
\sum_{j=1}^{m}\left|a_{i_j}\right|.
\label{eq:atomic-length}
\end{equation}

The term $|C|$ represents the cost of specifying how the selected atomic units are combined. This prevents the framework from treating decomposition as cost-free. Very fine-grained units may be individually compact but require long composition descriptions, whereas very coarse units may represent particular cases economically while providing little opportunity for reuse.

The atomic representation is not intended to replace the underlying evidence. Surface artefacts remain necessary for provenance, temporal context, auditability, and domain-specific detail. The atomic layer provides reusable reference structures through which related evidence can be organised and, where appropriate, composed.

The definition also establishes a stronger requirement than simple clustering. A set of evidence units grouped under one identifier is not automatically equivalent to an atomic representation of the form in Equation~\ref{eq:atomic-representation}. A complete atomic representation additionally requires an explicit or operational account of the units being composed and of the composition rule $C$. This distinction becomes important in the empirical evaluation, where persistent concepts are used as consolidation references but no composition grammar is inferred for them.

\subsection{Atomic and Domain Compression}
\label{subsec:atomic-domain-compression}

For a phenomenon $\phi \in D$, the Atomic Compression Ratio is defined as

\begin{equation}
\operatorname{ACR}(\phi)
=
\frac{\left|s(\phi)\right|}
     {\left|\alpha(\phi)\right|}.
\label{eq:acr}
\end{equation}

An $\operatorname{ACR}$ greater than $1$ indicates that the atomic representation is shorter under the selected measurement convention. A value close to $1$ indicates limited representational compression, while a value below $1$ indicates that the atomic representation, including its composition overhead, is more expensive than the corresponding surface representation.

The magnitude of an $\operatorname{ACR}$ has meaning only relative to the specified surface representation, atomic library, composition grammar, measurement unit, and fidelity requirement. It should therefore not be interpreted as an intrinsic property of the represented phenomenon.

For a distribution of phenomena in domain $D$, the Domain Compression Ratio is

\begin{equation}
\operatorname{DCR}(D)
=
\mathbb{E}_{\phi \sim D}
\left[
\operatorname{ACR}(\phi)
\right].
\label{eq:dcr}
\end{equation}

The $\operatorname{DCR}$ is an average over a stated distribution and representation scheme. It is therefore not an intrinsic constant of a domain. It may vary with the atomic library, the evidence-unit definition, the composition grammar, the fidelity criterion, and the cases included in the analysis.

The theoretical framework assumes that the atomic representation retains the information required for its intended use. Compactness without adequate fidelity does not constitute useful atomic compression. The empirical evaluation in Section~\ref{sec:evaluation} does not construct paired surface and atomic encodings and therefore does not directly estimate either $\operatorname{ACR}$ or $\operatorname{DCR}$. Instead, it measures evidence-to-concept consolidation, defined separately below.

\subsection{Evidence-to-Concept Consolidation}
\label{subsec:evidence-concept-consolidation}

The empirical evaluation measures evidence-to-concept consolidation as an operational property of a persistent reference layer. Let $\mathcal{U}$ be the set of observed evidence units and let $\mathcal{C}$ be the set of persistent reference concepts. Their cardinalities are

\begin{equation}
U = |\mathcal{U}|,
\qquad
C = |\mathcal{C}|.
\label{eq:evidence-concept-counts}
\end{equation}

The consolidation ratio is

\begin{equation}
R
=
\frac{U}{C},
\label{eq:consolidation-ratio}
\end{equation}

and the corresponding reduction is

\begin{equation}
\operatorname{Reduction}
=
1-\frac{C}{U}.
\label{eq:consolidation-reduction}
\end{equation}

A ratio of $20{:}1$ corresponds to a reduction of $95\%$. The quantity $R$ describes how many evidence units are organised, on average, around each persistent reference concept. It is a count-based structural measure. It does not by itself establish semantic equivalence among the associated evidence units, the quality of the resulting concept representation, the correctness of an automatically inferred merge, or the existence of a valid composition grammar.

For this reason, $R$ should not be identified directly with $\operatorname{ACR}$. The two quantities answer different questions. $\operatorname{ACR}$ compares the description length of a surface representation with that of an explicitly defined atomic representation. The consolidation ratio measures recurrence around reference concepts in an observed corpus. Consolidation can provide the structural substrate from which atomic representations may later be constructed, but a high consolidation ratio alone does not demonstrate semantic compression at fixed fidelity.

\paragraph{Evidence units.}
An evidence unit must be externally defined and reported with the result. In the software evaluation, two unit definitions are used:

\begin{enumerate}
    \item a message-level unit, in which one authored email is one evidence unit; and
    \item a deduplicated paragraph-level unit, in which messages are split on blank lines, short fragments are removed, and exact duplicate paragraphs are excluded.
\end{enumerate}

Because $R=U/C$, a finer unit definition can increase the reported ratio even when the underlying text is unchanged. The evidence-unit boundary is therefore part of the measurement rather than a neutral preprocessing choice. Results obtained under different unit definitions should be compared only with this dependence made explicit.

\paragraph{Closed-world concept model.}
Under a closed-world model, additional evidence is associated only with concepts already represented in a predefined identifier set. If $\mathcal{C}_{0}$ is the predefined concept set, evidence that cannot be linked to that set does not introduce a new concept. Depending on the corpus-construction rule, such evidence may therefore be excluded from the measurement or retained without expanding the denominator. In either case, the concept space is constrained before the new evidence is observed.

\paragraph{Open-world concept model.}
Under an open-world model, unlinked evidence is retained and may introduce additional concepts. In the principal empirical configuration, each unlinked reply thread contributes one additional reference concept. Let $\mathcal{C}_{\mathrm{linked}}$ denote concepts identified through existing issue keys and let $\mathcal{C}_{\mathrm{new}}$ denote concepts introduced from unlinked threads. Then

\begin{equation}
\mathcal{C}
=
\mathcal{C}_{\mathrm{linked}}
\cup
\mathcal{C}_{\mathrm{new}}.
\label{eq:open-world-concepts}
\end{equation}

Allowing both $U$ and $C$ to change makes the resulting ratio a measurement of the retained corpus under the stated identity rule rather than a construction based only on evidence that already conforms to an existing identifier system.

The open-world rule does not imply that every reply thread corresponds to a fully developed semantic concept. Thread identity is an operational proxy that allows previously unlinked evidence to remain within the observed concept space. The sensitivity of the result to alternative treatments of unlinked evidence is analysed explicitly in Section~5.5.

\paragraph{Cross-source corpus construction.}
Repository evidence and mailing-list evidence are processed through a common pipeline. Explicit issue-key references provide direct links to existing reference concepts, while reply-thread structure preserves discussion continuity. Automated messages, exact duplicates, empty or very short content, and evidence already represented in the repository corpus are identified during preprocessing. The same acquisition, filtering, unitisation, linking, and concept-identity rules are applied across the evaluated projects.

Cross-source construction should be distinguished from cross-source recurrence. The methodology permits evidence from heterogeneous sources to be represented under a common concept model, but it does not assume that the addition of a new source must produce a large increase in consolidation. The contribution of each evidence source is therefore analysed empirically rather than inferred from the architecture of the corpus.

\subsection{Compositional Compression Gain}
\label{subsec:compositional-compression-gain}

The Atomic Compression Ratio concerns the representation of an individual phenomenon. Atomic units provide an additional form of leverage when they can be reused within higher-order structures.

Suppose that two atomic units $a_1$ and $a_2$ compose into a higher-order unit

\begin{equation}
a_3 = C(a_1,a_2).
\label{eq:composed-atom}
\end{equation}

Let $\phi_3$ denote the surface phenomenon represented by $a_3$. The Compositional Compression Gain is defined as

\begin{equation}
\operatorname{CCG}(a_3)
=
\frac{\left|s(\phi_3)\right|}
     {|a_1|+|a_2|+|C|}.
\label{eq:ccg}
\end{equation}

The numerator measures the surface cost of representing the higher-order phenomenon directly. The denominator includes both the selected atomic units and the cost of specifying their composition. A high $\operatorname{CCG}$ therefore indicates that a larger structure can be represented economically through reusable components rather than reconstructed directly from its surface description.

The definition assumes that the composition preserves the information required for the intended use. A superficially compact combination that omits required information does not constitute a valid gain under the framework.

The empirical evaluation in Section~\ref{sec:evaluation} does not directly estimate $\operatorname{CCG}$. It measures evidence consolidation around persistent references. Consequently, no compositional gain is inferred solely from the consolidation ratios reported there.

\subsection{An Information-Theoretic Ceiling}
\label{subsec:compression-ceiling}

Compression gains are bounded by the information content of the represented domain. The Compounding Cascade therefore describes how compression ratios relate across successive representations, not a mechanism for unlimited reduction in description length.

\paragraph{Theorem (Bounded Amortised Compression).}
Let a domain $\mathcal{D}$ emit phenomena $X \sim P$ over an alphabet $\mathcal{X}$, with Shannon entropy

\begin{equation}
H(X)
=
-\sum_{x \in \mathcal{X}}P(x)\log_2 P(x).
\label{eq:shannon-entropy}
\end{equation}

Let $s(\cdot)$ and $a(\cdot)$ denote surface and atomic encodings, respectively, and assume that $a$ is uniquely decodable, $H(X)>0$, and $\mathbb{E}|s(X)|<\infty$. Then

\begin{equation}
\overline{R}
=
\frac{\mathbb{E}|s(X)|}
     {\mathbb{E}|a(X)|}
\leq
\frac{\mathbb{E}|s(X)|}
     {H(X)}
=
\rho_{\mathcal{D}}
<
\infty.
\label{eq:bounded-amortised-compression}
\end{equation}

\begin{proof}
Since $a$ is uniquely decodable, the Kraft--McMillan inequality gives

\begin{equation}
Z
=
\sum_{x \in \mathcal{X}}
2^{-|a(x)|}
\leq
1.
\end{equation}

Define

\begin{equation}
q(x)
=
\frac{2^{-|a(x)|}}{Z}.
\end{equation}

Then

\begin{align}
\mathbb{E}|a(X)|
&=
-\sum_x P(x)
\log_2\!\left(2^{-|a(x)|}\right) \\
&=
-\sum_x P(x)
\log_2\!\left(q(x)Z\right) \\
&=
H(X)
+
D(P\|q)
-
\log_2 Z.
\end{align}

By Gibbs' inequality, $D(P\|q)\geq0$, and because $Z\leq1$, $-\log_2 Z\geq0$. Hence,

\begin{equation}
\mathbb{E}|a(X)|
\geq
H(X).
\end{equation}

Dividing $\mathbb{E}|s(X)|$ by both sides yields the stated bound.
\end{proof}

The theorem establishes that the amortised gain achievable by a uniquely decodable atomic encoding remains constrained by the entropy of the represented source. Layered abstraction may expose progressively more useful structure, but it cannot remove information that must be preserved under the selected representation and fidelity conditions.

\paragraph{Compression at fixed fidelity.}
Where lossy representations are permitted, compactness must be considered together with the amount of distortion introduced. Given a distortion measure

\begin{equation}
d:
\mathcal{X}
\times
\widehat{\mathcal{X}}
\rightarrow
\mathbb{R}_{\geq0}
\end{equation}

and a distortion budget $D\geq0$, the rate--distortion function is

\begin{equation}
R(D)
=
\min_{\substack{p(\widehat{x}\mid x):\\
\mathbb{E}[d(X,\widehat{X})]\leq D}}
I(X;\widehat{X}).
\label{eq:rate-distortion}
\end{equation}

Where $R(D)>0$, a fidelity-indexed compression ratio may be defined as

\begin{equation}
\operatorname{ACR}(D)
=
\frac{\mathbb{E}|s(X)|}{R(D)}.
\label{eq:fidelity-acr}
\end{equation}

This formulation separates representational compactness from the distortion permitted by an application. It is part of the general theory and is not directly estimated in the current empirical evaluation.

\subsection{The Compounding Cascade}
\label{subsec:compounding-cascade}

Once an atomic representation has been formed, it may become the input to a higher abstraction layer. Functions can compose into modules, modules into services, and services into platforms. The paper refers to this recursive process as the \emph{Compounding Cascade}.

Let

\begin{equation}
L_0,L_1,\ldots,L_n
\label{eq:abstraction-stack}
\end{equation}

be a sequence of representational layers, where $L_0$ is the surface layer and each $L_i$ is defined through representations over $L_{i-1}$.

For a phenomenon $\phi$, let $\operatorname{repr}_i(\phi)$ denote its representation at layer $L_i$. Define the amortised Layer Compression Ratio as

\begin{equation}
\operatorname{LCR}_i
=
\frac{
\mathbb{E}_{\phi}
\left|\operatorname{repr}_{i-1}(\phi)\right|
}{
\mathbb{E}_{\phi}
\left|\operatorname{repr}_{i}(\phi)\right|
}.
\label{eq:lcr}
\end{equation}

\paragraph{Theorem (Compounding Cascade).}
The Total Compression Ratio of layer $L_n$ relative to $L_0$ is

\begin{equation}
\operatorname{TCR}_n
=
\frac{
\mathbb{E}_{\phi}
\left|\operatorname{repr}_{0}(\phi)\right|
}{
\mathbb{E}_{\phi}
\left|\operatorname{repr}_{n}(\phi)\right|
}
=
\prod_{i=1}^{n}\operatorname{LCR}_i.
\label{eq:tcr}
\end{equation}

\begin{proof}
The product telescopes:

\begin{align}
\prod_{i=1}^{n}\operatorname{LCR}_i
&=
\frac{\mathbb{E}|\operatorname{repr}_0|}
     {\mathbb{E}|\operatorname{repr}_1|}
\cdot
\frac{\mathbb{E}|\operatorname{repr}_1|}
     {\mathbb{E}|\operatorname{repr}_2|}
\cdots
\frac{\mathbb{E}|\operatorname{repr}_{n-1}|}
     {\mathbb{E}|\operatorname{repr}_n|} \\
&=
\frac{\mathbb{E}|\operatorname{repr}_0|}
     {\mathbb{E}|\operatorname{repr}_n|}
=
\operatorname{TCR}_n.
\end{align}
\end{proof}

Each $\operatorname{LCR}_i$ must be calculated as a ratio of aggregate or expected lengths. Multiplying averages of per-phenomenon ratios does not generally recover $\operatorname{TCR}_n$ because

\begin{equation}
\mathbb{E}\!\left[\frac{A}{B}\right]
\neq
\frac{\mathbb{E}[A]}{\mathbb{E}[B]}.
\end{equation}

\paragraph{Corollary (Equal layer ratios).}
If each of $n$ layers has the same compression ratio $r$, then

\begin{equation}
\operatorname{TCR}_n
=
r^n.
\label{eq:equal-layer-tcr}
\end{equation}

For example, if each of seven layers has ratio $5$, then

\begin{equation}
\operatorname{TCR}_7
=
5^7
=
78125.
\end{equation}

This numerical example illustrates the multiplicative identity. It is not an empirical estimate for the software corpora evaluated in Section~\ref{sec:evaluation}.

\paragraph{Corollary (Abstraction dividend).}
Adding a layer with compression ratio $\operatorname{LCR}_{n+1}$ gives

\begin{equation}
\operatorname{TCR}_{n+1}
=
\operatorname{TCR}_{n}
\operatorname{LCR}_{n+1}.
\label{eq:abstraction-dividend}
\end{equation}

A useful abstraction layer therefore multiplies the compression already achieved by the lower layers. Conversely, a layer that adds description or composition cost without sufficient representational gain can reduce the effectiveness of the stack. The Compounding Cascade should therefore not be interpreted as an assumption that every abstraction layer necessarily provides compression.

\subsection{Atomic Granularity and Quality}
\label{subsec:atomic-granularity-quality}

The effectiveness of an atomic library depends on both the granularity and quality of its units. Atomic units that are too fine-grained may create excessive composition cost. Units that are too coarse may represent individual cases compactly while providing little reuse or compositional flexibility.

For a corpus $\mathcal{X}$ and a candidate atomic library $\mathcal{L}$, let $L(\mathcal{L})$ denote the description length of the library and let $L(\mathcal{X}\mid\mathcal{L})$ denote the length of the corpus encoded through that library. An MDL-based granularity criterion is

\begin{equation}
\mathcal{L}^{\star}
=
\arg\min_{\mathcal{L}}
\left[
L(\mathcal{L})
+
L(\mathcal{X}\mid\mathcal{L})
\right].
\label{eq:mdl-granularity}
\end{equation}

A very fine library may contain compact atomic units but require long composition descriptions. A very coarse library may reduce composition cost while approaching the size and specificity of the corpus itself. The preferred granularity minimises the combined cost and is therefore dependent on the domain, corpus, task, and fidelity requirement.

\paragraph{Layer quality.}
Granularity alone does not ensure that the selected units capture the useful structure available at a layer. Let $\operatorname{LCR}_i^{\mathrm{opt}}$ be the optimal layer compression ratio under the selected representation and fidelity conditions, and let $\operatorname{LCR}_i^{\mathrm{actual}}$ be the ratio achieved by the selected atomic units. Define

\begin{equation}
q_i
=
\frac{
\operatorname{LCR}_i^{\mathrm{actual}}
}{
\operatorname{LCR}_i^{\mathrm{opt}}
},
\qquad
0<q_i\leq1.
\label{eq:quality-factor}
\end{equation}

The achieved total compression is then

\begin{equation}
\operatorname{TCR}_n^{\mathrm{actual}}
=
\left(
\prod_{i=1}^{n}q_i
\right)
\operatorname{TCR}_n^{\mathrm{opt}}.
\label{eq:quality-multiplier}
\end{equation}

Thus, representational losses at individual layers can compound multiplicatively. The quality factors are theoretical unless the corresponding optimum can be estimated under explicit representation and fidelity assumptions. No $q_i$ is estimated for the empirical corpora in Section~\ref{sec:evaluation}, and the observed consolidation ratios are not interpreted as estimates of $\operatorname{LCR}_i^{\mathrm{opt}}$ or of the distance from such an optimum.

This formulation also provides a connection to the soft-atom analysis in Section~\ref{sub:meaning-gap}. Context-dependent or weakly constrained representations may still encode useful structure, but to the extent that instability or invalid composition reduces the effective representation available at a layer, the resulting loss can be expressed conceptually through $q_i<1$. The present paper does not assign numerical quality factors to language-model representations.

\subsection{Application Protocol}
\label{subsec:application-protocol}

The Compression Calculus can be applied through the following protocol.

\paragraph{Step 1: Define the domain and evidence.}
Specify the domain $D$, the class of phenomena $\phi$, the available evidence sources, and the distribution over which any averages are calculated.

\paragraph{Step 2: Define the surface unit.}
Specify $s(\phi)$ and the unit used to measure $\left|s(\phi)\right|$. The unit boundary must be reproducible and externally motivated where possible.

\paragraph{Step 3: Define the atomic layer.}
Specify the atomic library $A_D$, the concept-identity rule, and, where composition is being analysed, the grammar $G_D$. Where the concept set may expand, define the open-world rule used to introduce new concepts.

\paragraph{Step 4: Measure compression or consolidation.}
For paired surface and atomic descriptions, calculate $\operatorname{ACR}$ or $\operatorname{DCR}$. For evidence associated with persistent reference concepts, calculate $U$, $C$, $R=U/C$, and the corresponding reduction. The resulting quantity must be identified explicitly as compression or consolidation.

\paragraph{Step 5: Report representation choices.}
Report the unit definition, concept model, filtering rules, deduplication rules, measurement basis, and fidelity assumptions where applicable. State whether the measurement is count-based or length-based. These choices form part of the result.

\paragraph{Step 6: Analyse composition where supported.}
Where explicit higher-order structures and composition rules are available, estimate $\operatorname{CCG}$ and the relevant layer ratios. Do not infer compositional gain solely from evidence-to-concept consolidation.

\paragraph{Step 7: Separate coverage from compression.}
For a usage distribution $P$ with frequencies ordered as

\begin{equation}
p_{(1)}
\geq
p_{(2)}
\geq
\cdots,
\end{equation}

the top-$k$ coverage is

\begin{equation}
\operatorname{Cov}_k(P)
=
\sum_{i=1}^{k}p_{(i)}.
\label{eq:coverage}
\end{equation}

Coverage measures usage concentration. Compression measures reduction in representation length or unit count relative to an explicitly defined alternative representation. A small vocabulary accounting for a large proportion of observations may therefore be consistent with an atomic representation, but high coverage alone does not establish a high compression ratio.

\paragraph{Step 8: State epistemic status.}
Every reported quantity should be identified as one of the following: a direct empirical measurement, an analytical estimate under stated assumptions, an illustrative numerical example, a theoretical bound, or a coverage statistic. Quantities belonging to different categories should not be treated as interchangeable evidence for the same claim.

\paragraph{Empirical instantiation.}
Section~\ref{sec:evaluation} applies the protocol to multi-source software corpora. Repository artefacts and mailing-list fragments form the evidence set, JIRA work items and open-world reply-thread concepts form the persistent reference layer, and the ratio $U/C$ measures consolidation. Message-level and deduplicated paragraph-level units are reported separately, and the same corpus-construction procedure is applied to HBase, Hadoop, and Maven.

The empirical evaluation therefore tests the consolidation component of the framework. It does not directly measure $\operatorname{ACR}$, $\operatorname{DCR}$, semantic fidelity, automatic atom discovery, $\operatorname{CCG}$, or multi-layer compression. Those claims require separate experimental procedures.

\subsection{Cross-Domain Cascade Estimates}
\label{subsec:cross-domain-cascades}

The protocol above can also be used analytically to illustrate how the Compounding Cascade would operate when explicit abstraction layers and approximate representation sizes are supplied. This subsection applies that procedure to four domains: software engineering, customer service, medicine, and music. These domains are used because they provide comparatively interpretable abstraction hierarchies and can be connected to the independent empirical literature discussed later in Section~5.7.

The quantities in this subsection are \emph{analytical estimates under stated assumptions}. They are not measurements obtained from the Section~\ref{sec:evaluation} corpora, and they should not be interpreted as empirically established domain compression ratios. Their purpose is to demonstrate how the formalism can be instantiated and to make the assumptions behind a multi-layer calculation explicit. Appendix~A provides the corresponding worked examples.

Table~\ref{tab:cascade-estimates} summarises the four illustrative cascades.

\begin{center}
\footnotesize
\setlength{\tabcolsep}{3pt}
\renewcommand{\arraystretch}{1.15}

\textbf{Illustrative cross-domain cascade estimates.}\\
\emph{All values are analytical estimates under stated assumptions rather than empirical measurements.}

\vspace{0.5em}

\begin{tabular}{
p{2.0cm}
p{5.2cm}
p{3.0cm}
p{1.8cm}
p{2.0cm}
}
\hline
\textbf{Domain} &
\textbf{Illustrative abstraction layers} &
\textbf{Approx.\ layer ratios} &
\textbf{Approx.\ TCR} &
\textbf{Status} \\
\hline

Software engineering &
Function $\rightarrow$ module $\rightarrow$ service $\rightarrow$ platform &
$30\times$, $12\times$, $8\times$, $20\times$ &
$57600\times$ &
Analytical estimate \\

Customer service &
Utterance $\rightarrow$ intent $\rightarrow$ journey $\rightarrow$ insight &
$47\times$, $16\times$, $200\times$ &
$\sim150000\times$ &
Analytical estimate \\

Medicine &
Clinical data $\rightarrow$ diagnosis $\rightarrow$ protocol $\rightarrow$ pathway &
$160\times$, $250\times$, $10\times$ &
$\sim400000\times$ &
Analytical estimate \\

Music &
Notes $\rightarrow$ chords $\rightarrow$ progressions $\rightarrow$ sections $\rightarrow$ form &
$4\times$, $8\times$, $4\times$, $6\times$ &
$768\times$ &
Analytical estimate \\

\hline
\end{tabular}

\end{center}
The software estimate illustrates the calculation most directly. Under an assumption of approximately $30$ lines for the surface implementation represented by an average reusable function, a function call provides an approximate first-layer ratio of $30\times$. Assuming approximately $12$ such functions per module, eight modules per service, and twenty services per platform gives

\begin{equation}
30
\times
12
\times
8
\times
20
=
57600.
\end{equation}

The particular values are assumptions used to expose the calculation rather than population estimates for software systems. Appendix~A also considers a more concrete approximately $25$-line single-function example. The distinction between the worked function example and the average value used in the illustrative cascade is retained explicitly.

The quality factor in Equation~\ref{eq:quality-multiplier} gives this illustration an additional interpretation. If a four-layer software cascade had $q_i=0.5$ at each layer, then

\begin{equation}
57600
\times
(0.5)^4
=
3600.
\end{equation}

The example demonstrates how representational quality can affect a layered estimate multiplicatively. It does not imply that $q_i=0.5$ is observed in software systems, nor that $57600\times$ or $3600\times$ has been measured empirically.

The customer-service estimate uses an approximately $47\times$ utterance-to-intent representation, followed by an assumed $16\times$ reduction from recurring intent sequences to journey patterns and an approximately $200\times$ reduction from recurring journeys to an operational-insight representation. Under these assumptions, the product is approximately $150000\times$.

The medical illustration uses three assumed levels: clinical observations represented through a diagnostic pattern, diagnosis represented through a treatment protocol, and related protocols represented through a broader clinical pathway. The corresponding illustrative ratios of approximately $160\times$, $250\times$, and $10\times$ give a total of approximately $400000\times$. These quantities should not be interpreted as measurements of clinical information loss or diagnostic performance.

The music illustration uses a more modest four-layer hierarchy in which notes are represented through chords, chords through progressions, progressions through sections, and sections through a larger musical form. Approximate layer ratios of $4\times$, $8\times$, $4\times$, and $6\times$ yield an illustrative total of $768\times$.

These cascades are deliberately restricted to the four domains above. They do not establish that every domain admits the same abstraction depth, that the selected layers are unique, or that the resulting ratios would survive empirical measurement under a fixed fidelity requirement. In particular, none of the four estimates has been checked against an empirical estimate of the corresponding domain entropy. Establishing such an entropy estimate is a prerequisite for treating the illustrative values as candidate empirical compression ratios rather than demonstrations of the calculus. The entropy-bounded prediction in Section~\ref{sub:predictions} provides a direct route for testing this condition.

The cross-domain estimates therefore serve a methodological rather than confirmatory role. They demonstrate how the same definitions can be instantiated under different representational assumptions, while the explicit status labels prevent analytical examples from being conflated with the empirical consolidation measurements reported in Section~\ref{sec:evaluation}.

\section{Experimental Evaluation}
\label{sec:evaluation}

This evaluation measures evidence-to-concept consolidation across three multi-source software project corpora: HBase, Hadoop, and Maven. The principal configuration uses an open-world, thread-based concept model and evaluates both message-level and deduplicated paragraph-level evidence. At paragraph granularity, the target ratio of $20{:}1$, corresponding to a $95\%$ reduction, is reached in the two larger projects, HBase and Hadoop, but not in the smaller Maven corpus. The target is not reached at message granularity in any of the three projects. The empirical result is therefore conditional on corpus scale, evidence-unit definition, and concept-identity rule.

Let $U$ denote the number of evidence units in a corpus and $C$ the number of persistent reference concepts. As defined in Section~\ref{subsec:evidence-concept-consolidation}, the consolidation ratio and corresponding reduction are

\[
R = \frac{U}{C},
\qquad
\mathrm{Reduction} = 1 - \frac{C}{U}.
\]

The quantity $R$ is a count-based consolidation measure. It does not constitute a direct estimate of the Atomic Compression Ratio, semantic fidelity, or compositional gain. The evaluation instead tests whether substantial collections of heterogeneous software evidence recur around a smaller set of persistent reference structures under explicitly defined unit and concept models.

\subsection{Datasets and Corpus Construction}
\label{subsec:evaluation-data}

The evaluation combines software repository evidence from SEOSS-33 with public Apache mailing-list archives. SEOSS-33 is a collection of 33 open-source software projects containing issue-tracking artefacts, version-control artefacts, metadata, and trace links \citep{rath2019seoss}. Across the project databases, the extraction used in the released evaluation contains $2434469$ evidence units associated with $278563$ reference work items. JIRA issue keys provide the principal persistent reference identifiers.

The SEOSS-33 dataset is publicly available from Harvard Dataverse at

\begin{center}
\url{https://doi.org/10.7910/DVN/PDDZ4Q}
\end{center}

The repository evidence was extended with developer and user mailing-list archives obtained from the official Apache archive infrastructure:

\begin{center}
\url{https://lists.apache.org/}
\end{center}

Three Apache projects are evaluated. The HBase mailing-list collection covers January 2009 to November 2017 and contains $109179$ unique retrieved messages. The Hadoop collection contains $134309$ retrieved messages across 614 monthly archive files. The Maven collection contains $20285$ retrieved messages, of which 338, approximately $2\%$, were identified as automated traffic.

The mailing-list evidence contains technical discussion, proposed solutions, design decisions, implementation details, user reports, and exchanges between project participants. Messages are associated with persistent project concepts through explicit JIRA issue-key references and reply-thread structure.

Automated messages, exact duplicates, empty or very short content, and material already represented in the repository evidence are identified during preprocessing. The same acquisition, filtering, linking, unitisation, concept-identification, and measurement procedures are applied across HBase, Hadoop, and Maven.

For reproducibility, the download manifests record the source URL, byte size, and SHA-256 checksum of the retrieved archive files.

\subsection{Evidence Units and Concept Identification}
\label{subsec:evidence-units}

The evaluation uses two definitions of an evidence unit. At message level, each authored email is treated as one evidence unit. This boundary corresponds to an artefact produced and sent by a project participant.

At paragraph level, messages are divided on blank lines. Paragraphs shorter than 20 characters are excluded, and exact duplicate paragraphs are removed before the consolidation ratio is calculated. This representation captures distinct technical statements, observations, explanations, proposed actions, and decisions that may occur within a single message.

Repository evidence and mailing-list evidence are associated with persistent reference concepts primarily through JIRA issue keys. Where a message contains an explicit issue key, it is linked directly to the corresponding work item. Reply-thread information is used to preserve discussion continuity where an identifier occurs in only part of an exchange.

The evidence-unit definition is part of the measurement. Moving from complete messages to paragraphs increases $U$ because one authored artefact may contain several qualifying evidence units. Consequently, the paragraph-level ratio cannot be interpreted simply as a more precise estimate of the message-level ratio. The two measurements answer the same consolidation question under different unit boundaries.

The procedure creates a combined evidence base in which formal repository records and less structured developer communication can be represented through the same reference-concept framework. It does not assume that every reference concept constitutes a fully formed atomic unit as defined theoretically in Sections~\ref{sub:atomic-units-abstraction} and~\ref{sub:meaning-gap}.

\subsection{Open-World Concept Model}
\label{subsec:open-world}

A central methodological component of the evaluation is the open-world treatment of concept identity.

In a closed-world measurement, newly collected evidence contributes only when it can be associated with a concept already represented in the repository identifier set. Evidence that cannot be linked to a known issue does not introduce an additional concept. The concept space is therefore constrained before the additional evidence source is incorporated.

This can affect the measured ratio. Linked evidence increases the number of evidence units, $U$, while unlinked discussions do not expand $C$. A closed-world measurement can consequently reflect the boundaries of the existing identifier system rather than the structure of all retained evidence.

The open-world model retains evidence that cannot be linked to a pre-existing JIRA issue. In the principal configuration, each unlinked reply thread is treated as one additional reference concept. This permits both $U$ and $C$ to change as mailing-list evidence is incorporated.

A mailing-list discussion can therefore contribute to the measurement even where its participants do not explicitly refer to a JIRA issue. The resulting ratio is calculated over the retained corpus rather than only over the subset already compatible with the repository identifier scheme.

Reply-thread identity is an operational proxy rather than an assertion that every thread corresponds to a semantically complete atomic unit. Alternative assumptions about unlinked evidence produce different values of $C$. The sensitivity of the result to this choice is examined in Section~\ref{subsec:boundary-conditions}.

\subsection{Primary Multi-Corpus Results}
\label{subsec:primary-results}

The primary experiment applies the same open-world, thread-level concept model to HBase, Hadoop, and Maven. Each project is evaluated using both message-level and deduplicated paragraph-level evidence.

Table~\ref{tab:primary-results} reports all six principal measurements. The ratios are shown to three decimal places in the table so that the near-equality of the HBase and Hadoop message-level results remains visible rather than being hidden by rounding.

\begin{table*}[t]
\centering
\small
\begin{tabular}{llrrrrl}
\hline
\textbf{Project} &
\textbf{Evidence unit} &
$\mathbf{U}$ &
$\mathbf{C}$ &
$\mathbf{U{:}C}$ &
\textbf{Reduction} &
\textbf{20:1 reached} \\
\hline
HBase &
Message &
$367590$ &
$29026$ &
$12.664{:}1$ &
$92.1\%$ &
no \\

HBase &
Deduplicated paragraph &
$664480$ &
$29026$ &
$22.893{:}1$ &
$95.6\%$ &
yes \\

Hadoop &
Message &
$597232$ &
$47187$ &
$12.657{:}1$ &
$92.1\%$ &
no \\

Hadoop &
Deduplicated paragraph &
$949131$ &
$47187$ &
$20.114{:}1$ &
$95.0\%$ &
yes \\

Maven &
Message &
$45689$ &
$8571$ &
$5.331{:}1$ &
$81.2\%$ &
no \\

Maven &
Deduplicated paragraph &
$98004$ &
$8571$ &
$11.434{:}1$ &
$91.3\%$ &
no \\
\hline
\end{tabular}
\caption{Open-world, thread-level consolidation results across the three evaluated software projects.}
\label{tab:primary-results}
\end{table*}

For HBase, the message-level corpus contains $367590$ evidence units associated with $29026$ concepts, producing a ratio of $12.66{:}1$ and a reduction of $92.1\%$. Following paragraph extraction and exact deduplication, the corpus contains $664480$ evidence units associated with the same open-world concept set. The resulting ratio is $22.89{:}1$, corresponding to a reduction of $95.6\%$.

For Hadoop, the message-level corpus contains $597232$ evidence units associated with $47187$ concepts, producing a ratio of $12.66{:}1$ and a reduction of $92.1\%$. At deduplicated paragraph level, the corpus contains $949131$ evidence units associated with $47187$ concepts, producing a ratio of $20.11{:}1$ and a reduction of $95.0\%$.

The HBase and Hadoop message-level ratios both round to $12.66{:}1$ at two decimal places. This equality is coincidental: before rounding, the ratios are $12.664{:}1$ and $12.657{:}1$, respectively.

For Maven, the message-level corpus contains $45689$ evidence units associated with $8571$ concepts, giving a ratio of $5.33{:}1$ and a reduction of $81.2\%$. At deduplicated paragraph level, the corpus contains $98004$ evidence units associated with the same concept set, producing a ratio of $11.43{:}1$ and a reduction of $91.3\%$.

The target is therefore reached at deduplicated paragraph granularity in the two larger projects, HBase and Hadoop, under the thread-level open-world model. It is not reached in the third and smallest project, Maven, and it is not reached at message granularity in any of the three projects. The result demonstrates replication across two projects while the Maven result establishes an empirical boundary on the scale and recurrence conditions under which the target is observed.

\subsection{Boundary Conditions}
\label{subsec:boundary-conditions}

The primary results vary substantially across projects and measurement configurations. These differences provide information about the conditions under which consolidation arises and prevent the $20{:}1$ result from being interpreted as a domain-wide constant.

Maven enters the multi-source experiment with substantially less recurrence in its repository evidence. Its repository-only base ratio is $5.85{:}1$, compared with $15.60{:}1$ for HBase and $14.26{:}1$ for Hadoop. The addition of mailing-list evidence therefore begins from different corpus structures. Under the current measurement, the target is observed at the scale and recurrence structure of HBase and Hadoop but not Maven. Corpus scale and within-project recurrence are consequently conditions on the empirical result alongside the evidence-unit definition.

The contribution of the second source is also smaller than the overall multi-source ratios might suggest. Adding the complete mailing-list collections changes the corresponding closed-world ratios by only $+1.39$ for HBase, $+0.39$ for Hadoop, and $+0.53$ for Maven. Cross-source recurrence is therefore not the principal driver of the reported consolidation. Most of the effect arises from recurrence already present within the project evidence and from the selected evidence-unit boundary. Maven's mailing-list collection is approximately $98\%$ human-authored, so its lower ratio is not explained by an unusually large proportion of automated traffic.

The treatment of unlinked evidence provides a second boundary. For HBase at paragraph granularity, the principal thread-level open-world model produces $22.89{:}1$. Treating every unlinked message as a separate additional concept reduces the ratio to $11.10{:}1$. These two rules provide an operational bracket rather than a unique estimate of semantic concept identity. The reported value is the thread-level endpoint of that bracket.

The open-world model is also more conservative than the corresponding closed-world result. For HBase at paragraph granularity, the closed-world ratio is $28.05{:}1$, whereas the reported open-world value is $22.89{:}1$. Retaining unlinked evidence while allowing it to expand the concept count therefore lowers rather than inflates the headline ratio.

Table~\ref{tab:boundary-comparisons} places the principal result alongside selected internal sensitivity and contrasting-corpus measurements.

\begin{table}[t]
\centering
\small
\begin{tabular}{lll}
\hline
\textbf{Case} &
\textbf{Ratio} &
\textbf{Role} \\
\hline
HBase, paragraph, closed world &
$28.05{:}1$ &
Identifier-constrained comparison \\

HBase, paragraph, open world, threads &
$22.89{:}1$ &
Principal configuration \\

HBase, paragraph, open world, messages &
$11.10{:}1$ &
Unlinked-concept sensitivity \\

Maven, repository base &
$5.85{:}1$ &
Scale/recurrence contrast \\

MeetingBank &
$1.06{:}1$ &
Low-recurrence contrast \\

MedMentions, mention level &
$10.15{:}1$ &
Curated concept identity \\

MedMentions, document level &
$5.39{:}1$ &
Curated concept identity \\
\hline
\end{tabular}
\caption{Selected boundary and contrasting-corpus measurements}
\label{tab:boundary-comparisons}
\end{table}

MeetingBank provides a particularly strong negative case. Its released measurement contains $6892$ evidence units associated with $6496$ agenda-item concepts, producing a ratio of $1.06{:}1$ and a reduction of only $5.7\%$. Natural transcript evidence therefore does not necessarily exhibit substantial recurrence around a smaller concept set.

MedMentions provides a different contrast because concept identity is established through curator-assigned UMLS concepts rather than software work-item identifiers. Its released measurements yield $10.15{:}1$ at mention level and $5.39{:}1$ at document level. Within-document repetition accounts for approximately $52\%$ of the reduction in that corpus. The comparison should not be interpreted as evidence that the software concepts are semantically finer. UMLS concepts represent semantic entities, whereas a JIRA issue is a work item that may legitimately aggregate heterogeneous evidence concerning a broader technical activity. In addition, the paragraph-level software measurements remove exact duplicate paragraphs before calculating the ratio, whereas the MedMentions analysis explicitly quantifies within-document repetition.

The theoretical criteria introduced in Section~\ref{sub:meaning-gap} create a further boundary on interpretation. JIRA issue keys provide persistent identity. Reply threads provide a weaker operational form of continuity. Neither the JIRA concepts nor the thread-level proxies are shown in the present evaluation to possess an explicit composition grammar, and compositional stability is not measured. The experiment therefore measures consolidation around persistent reference structures rather than the presence of fully formed atomic units satisfying all three theoretical properties.

Finally, no layer-quality factor $q_i$ is estimated for these corpora. The measured consolidation ratios are direct corpus quantities and are not claimed to approximate an optimal atomic representation or the quality ratio defined in Section~\ref{subsec:atomic-granularity-quality}.

\subsection{Interpretation}
\label{subsec:evaluation-interpretation}

The results demonstrate that substantial evidence-to-concept consolidation can occur in large software-engineering corpora, but they also show that the magnitude of the effect depends on how the corpus and its representational units are defined.

In HBase and Hadoop, many repository records, commits, comments, and developer communications recur around the same persistent work items. At deduplicated paragraph granularity, this produces consolidation above or approximately equal to the $20{:}1$ target. Maven contains less recurrence in its repository base and remains substantially below the target under the same procedure.

The unit-definition effect is equally important. None of the three projects reaches $20{:}1$ when complete messages are treated as evidence units. The higher paragraph-level ratios therefore reflect the fact that authored messages can contain several distinct evidence units associated with the same persistent reference. This is a representational result rather than a neutral preprocessing effect.

The multi-source analysis provides a complementary negative finding. Incorporating a long-running communication archive adds relatively little consolidation beyond that already visible in the repository evidence. The principal value of the cross-source procedure is therefore not that evidence from a second source necessarily creates substantial additional recurrence. It is that heterogeneous evidence can be retained, linked, and measured under a common open-world concept model without discarding material that does not already conform to the repository identifier system.

Consolidation should not be interpreted as deletion or semantic interchangeability. Individual messages, comments, commits, and issue records preserve differences in wording, purpose, provenance, date, and technical detail. Persistent reference concepts provide an organisational structure through which this evidence can be connected while the original artefacts remain available.

The evaluation consequently provides empirical evidence for a structural prerequisite of the broader atomic-units framework: recurrent evidence can, in some corpora, be organised around a substantially smaller reference layer. It does not establish semantic atom discovery, fidelity, retrieval performance, compositional gain, or higher-order cascade compression.

\subsection{Empirical Anchors from Published Literature}
\label{subsec:empirical-anchors}

The consolidation measurements above can be considered alongside independently published empirical results from other settings. None of the studies in this subsection was designed to evaluate the Compression Calculus. The comparison is therefore used to identify compatible empirical regularities and relevant constraints rather than to treat third-party results as validations of the framework.

A critical distinction is required between compression-style measurements and coverage or concentration measurements. File deduplication and cross-entropy directly measure forms of representational redundancy or predictability. By contrast, the proportion of observations assigned to a small set of intents, clinical codes, functionalities, or harmonic categories measures usage concentration. High coverage may be consistent with a reusable atomic vocabulary, but it does not establish a compression ratio under the definitions of Section~\ref{subsec:application-protocol}.

\begin{center}
\scriptsize
\setlength{\tabcolsep}{3pt}
\renewcommand{\arraystretch}{1.15}

\textbf{Published empirical anchors relevant to compression, recurrence, and usage concentration.}\\
\emph{Coverage measurements constrain but do not confirm the compression claims of the present framework.}

\vspace{0.5em}

\begin{tabular}{
p{0.25\linewidth}
p{0.68\linewidth}
}
\hline
\textbf{Domain and study} &
\textbf{Published finding and evidential status} \\
\hline

\textbf{Software}\\
Lopes et al.~\citep{lopes2017dejavu} &
More than 428 million files from 4.5 million non-fork GitHub projects were reduced to approximately 85 million unique files, corresponding to approximately $5{:}1$ at file level. Only $6\%$ of JavaScript files were distinct, corresponding to approximately $17{:}1$. \emph{Compression-style evidence.} \\

\textbf{Software}\\
Hindle et al.~\citep{hindle2012naturalness} &
$n$-gram cross-entropy for Java source code fell below 2 bits per token in some project corpora, substantially below the corresponding English corpus at higher $n$-gram orders. \emph{Compression-style evidence.} \\

\textbf{Software}\\
Svajlenko and Roy~\citep{svajlenko2015bigclonebench} &
BigCloneBench contained approximately 8 million validated clone pairs spanning 43 targeted functionalities. \emph{Coverage/concentration evidence.} \\

\textbf{Customer service}\\
Casanueva et al.~\citep{casanueva2020intent} &
BANKING77 contains $13083$ customer-service queries labelled with 77 banking intents. \emph{Coverage evidence.} \\

\textbf{Customer service}\\
Larson et al.~\citep{larson2019evaluation} &
CLINC150 contains 150 in-scope intents; the full dataset contains $23700$ examples, including out-of-scope examples. \emph{Coverage evidence.} \\

\textbf{Customer service}\\
FitzGerald et al.~\citep{fitzgerald2023massive} &
MASSIVE contains approximately 1 million parallel labelled utterances spanning 60 intents, 18 domains, and 51 languages. \emph{Coverage evidence.} \\

\textbf{Medicine}\\
Fenton et al.~\citep{fenton2025icd} &
Of $71932$ available ICD-10-CM codes, $32108$, or $44.6\%$, were used in the analysed 2019 ambulatory dataset; $33.3\%$ of the codes used were unspecified. \emph{Coverage evidence.} \\

\textbf{Medicine}\\
Arvig et al.~\citep{arvig2022chief} &
Across $223612$ adult non-trauma emergency-department visits, the six most frequent reported chief-complaint categories accounted for $70\%$ of visits. \emph{Coverage evidence.} \\

\textbf{Music}\\
Burgoyne~\citep{burgoyne2011stochastic} &
Five chord types accounted for $85.7\%$ of chords in the analysed McGill Billboard material. \emph{Coverage evidence.} \\

\textbf{Music}\\
de Clercq and Temperley~\citep{declercq2011corpus} &
The five most frequent chromatic roots accounted for $87\%$ of harmonies in the analysed rock corpus. \emph{Coverage evidence.} \\

\hline
\end{tabular}

\end{center}

The software results from Lopes et al. and Hindle et al. are the closest to direct compression-style evidence. File deduplication explicitly compares a surface population with a reduced set of unique files, while cross-entropy quantifies predictive regularity in source code.

The remaining results measure concentration rather than compression. BANKING77, CLINC150, and MASSIVE demonstrate that large sets of utterances can be classified using comparatively small intent vocabularies, but the number of utterances per intent is not an Atomic Compression Ratio. BigCloneBench demonstrates recurrence around a limited set of targeted functionalities, but the ratio of clone pairs to functionality classes is likewise not a valid description-length compression measurement.

The medical and musical findings provide similar constraints. A minority of available ICD-10-CM codes accounts for all codes observed in the ambulatory dataset, and a small set of emergency-department chief complaints accounts for a large fraction of visits. In the musical corpora, a small set of chord types or chromatic roots accounts for a large proportion of harmonic events. These distributions are consistent with strong usage concentration, but the Compression Calculus requires explicit surface and atomic representation lengths before they can be interpreted as compression.

The published literature therefore supports a more limited conclusion than a direct cross-domain replication claim. Recurrent concentration around comparatively small vocabularies is observed in several domains, while genuine compression-style measurements in software demonstrate substantial redundancy and predictability. The coverage results constrain the plausible structure of atomic representations but do not, by themselves, confirm the consolidation ratios or analytical cascade estimates reported elsewhere in this paper.

\subsection{Reproducibility and Data Availability}
\label{subsec:data-availability}

The empirical evaluation is based on publicly accessible source datasets and archives. SEOSS-33 is available through Harvard Dataverse, and the mailing-list evidence was obtained from the public Apache archive infrastructure described in Section~\ref{subsec:evaluation-data}.

To support reproducibility without requiring disclosure of proprietary implementation code, the accompanying research materials provide the information necessary to reconstruct and verify the reported measurements. These materials include the source-data acquisition manifests, filtering criteria, deduplication rules, evidence-unit definitions, issue-key linking rules, open-world concept definitions, and the intermediate and aggregate measurement outputs used in the analysis.

The primary multi-corpus measurements are reported in
\texttt{reports/multicorpus.md}. Project-level measurement outputs are provided under
\texttt{output/multicorpus/}, while the complete SEOSS project measurements are recorded in
\texttt{output/seoss\_all\_projects.json}. The contrasting-corpus analyses used for the boundary-condition evaluation are documented in
\texttt{reports/meetingbank.md} and
\texttt{reports/medmentions.md}.

The mailing-list acquisition manifests record the source URL, byte size, and SHA-256 checksum of each retrieved archive. The released measurement outputs expose the evidence counts, concept counts, consolidation ratios, reductions, source-specific contributions, and sensitivity analyses reported in the paper. These outputs allow the numerical results to be checked independently against the stated definitions and measurement procedure.

The production software used internally to execute parts of the data-processing pipeline is not released because it forms part of a proprietary software system. Reproducibility is therefore provided at the level of the experimental procedure, input sources, transformation rules, intermediate measurements, and reported outputs rather than through publication of the production codebase. A platform-independent description of the procedure is provided in Sections~\ref{subsec:evidence-units}--\ref{subsec:boundary-conditions} so that an independent implementation can reproduce the same measurement protocol.
\section{Discussion}
\label{sec:discussion}

The empirical evaluation provides conditional support for the central proposition of the atomic-units framework: substantial collections of software evidence can be organised around a considerably smaller set of persistent reference concepts. Under the thread-level open-world model and at deduplicated paragraph granularity, HBase reaches a consolidation ratio of $22.89{:}1$, corresponding to a reduction of $95.6\%$, while Hadoop reaches $20.11{:}1$, corresponding to $95.0\%$. Maven reaches $11.43{:}1$, corresponding to $91.3\%$, and therefore remains below the $20{:}1$ target. At message granularity, none of the three projects reaches that target. The empirical claim is consequently not that a fixed consolidation ratio applies generally, but that consolidation of this magnitude is observed in the two larger projects under the stated unit definition and concept model.

The boundary-condition analysis in Section~5.5 is important to the interpretation of this result. The three projects enter the evaluation with different levels of within-project recurrence, and the inclusion of mailing-list evidence changes the corresponding closed-world ratios only modestly. The principal source of the measured consolidation is therefore not recurrence introduced specifically by combining sources. Instead, corpus scale, recurrence already present within the software evidence, the evidence-unit boundary, and the concept-identity rule are all material components of the result. The multi-source contribution lies in demonstrating how heterogeneous evidence can be retained and organised under a common open-world concept model, including evidence that cannot be mapped to a pre-existing identifier.

The open-world model also makes the reported consolidation more conservative than the corresponding closed-world configuration because unlinked evidence is permitted to increase the concept count rather than being excluded. At the same time, the precise treatment of unlinked discussions remains consequential. Thread-level and message-level assignments provide different estimates of the concept set, and the reported thread-level configuration should therefore be understood as one explicitly defined operational model rather than as a unique ground-truth segmentation of the corpus.

A further distinction concerns the status of the reference concepts themselves. JIRA issue keys provide persistent identities around which heterogeneous evidence can accumulate. Reply-thread concepts provide a weaker operational form of continuity for discussions that lack such identifiers. Neither class of concept is shown in this evaluation to possess an explicit composition grammar, and the evaluation does not establish compositional stability for the resulting representations. The measurements therefore demonstrate evidence-to-concept consolidation around persistent references rather than the existence of fully developed atomic units satisfying every theoretical property introduced in Section~3.6.

This distinction is important because consolidation does not imply that the original evidence can be discarded or that evidence associated with the same concept is semantically interchangeable. Repository records, comments, commits, and mailing-list messages retain differences in wording, provenance, purpose, temporal position, and technical detail. The persistent concept instead provides a reference point through which distributed evidence can be connected and navigated while the original artefacts remain available.

The following subsections consider the implications of these findings for knowledge representation, LLM-based systems, dynamic knowledge composition, evolving atomic libraries, and the empirical predictions that follow from the framework.

\subsection{Implications for Knowledge Representation}
\label{sub:implications}

The evaluation supports the view that knowledge representation is partly a problem of selecting an appropriate representational level. Raw documents, messages, comments, commits, and issue records preserve valuable context, but information concerning the same subject may be distributed across many such artefacts. At the other extreme, tokens and isolated textual fragments provide useful computational units but do not necessarily correspond to persistent concepts.

The evaluated corpora illustrate an intermediate representational level. A single work item may be reflected in an issue description, several comments, related commits, technical discussions, implementation decisions, and follow-up messages. These artefacts remain distinct evidence units, but they can be organised around a shared reference concept. The empirical result therefore supports an evidence-and-concept model in which source artefacts and conceptual references coexist rather than one replacing the other.

In such an architecture, original evidence provides provenance, temporal context, and domain-specific detail. Persistent concepts provide stable points through which this evidence can be indexed and related. A practical atomic layer may consequently associate a concept identity with descriptive information, relations to other concepts, provenance links, temporal information, and references to the evidence from which the representation is derived.

This interpretation is relevant to domains in which the same subject is repeatedly expressed through different operational systems. In software engineering, an implementation problem may appear in an issue tracker, commit history, review discussion, and mailing-list exchange. In legal settings, a concept may be distributed across legislation, case law, administrative guidance, and contractual provisions. In scientific environments, a concept may recur across publications, datasets, annotations, and experimental records. These examples do not establish equivalent compression ratios across domains, but they illustrate the broader representational problem addressed by the framework.

The value of an atomic layer lies in making recurring conceptual references explicit. Without such a layer, relationships among separate surface artefacts must be reconstructed when the evidence is retrieved or interpreted. With a persistent reference layer, related evidence can be connected while retaining its original source and context.

The open-world analysis further suggests that the concept space should not be restricted to identifiers already present in a source system. JIRA issue keys provide reproducible anchors, but technical discussions can concern meaningful subjects without mentioning an issue identifier. Retaining these discussions and allowing them to introduce candidate concepts prevents the representation from equating the boundaries of an existing administrative system with the boundaries of the observed knowledge.

The broader methodological implication is that a knowledge architecture should be capable of extending its concept space when new evidence cannot be represented adequately through the existing vocabulary. A closed system may force new evidence into established categories or discard it from the conceptual representation. An open atomic system should instead support both consolidation onto existing concepts and controlled introduction of new concepts when the evidence requires them.

This motivates a layered view of representation. Original evidence occupies the surface layer. Persistent concepts connect related evidence at an intermediate level. Relations among concepts and recurring configurations can support subsequent abstraction layers. The empirical evaluation establishes only the consolidation component of this architecture, but the broader layered structure is consistent with the Compounding Cascade developed in the theoretical framework.

This approach connects naturally with knowledge graphs, ontologies, traceability systems, semantic memory, and neuro-symbolic architectures. Atomic units need not correspond only to ontology classes. Depending on the domain, candidate units may include work items, requirements, decisions, procedures, clauses, risks, events, or other persistent reference objects. Their role is to provide stable structures through which distributed evidence can be organised and subsequently composed.

\subsection{Implications for LLM-Based Systems}
\label{sub:implicationsforllm}

The findings also have implications for retrieval-augmented generation, agentic systems, and long-term semantic memory. Such systems commonly operate over two principal representational levels. Language models process token sequences, while external retrieval systems typically index documents, passages, messages, or fixed-size chunks.

Neither level necessarily corresponds to the persistent concepts around which evidence accumulates. A document may contain several concepts, while evidence concerning one concept may be distributed across multiple documents and sources. Fixed-size chunks preserve local text but may divide material belonging to a shared conceptual history. Token-level processing can interpret such evidence dynamically, but it does not by itself provide a persistent external identity for the underlying concept.

The software corpora provide a concrete instance of this representational problem. Information concerning one work item can appear across issue records, commits, comments, and developer discussions. A retrieval system that treats these artefacts as unrelated units may retrieve only part of the relevant evidence or return several fragments without representing why they belong together.

An atom-centred architecture would introduce an intermediate layer between the evidence store and the language model. Original evidence would remain available, but it would also be associated with persistent concepts. Retrieval could therefore proceed through both textual similarity and conceptual identity. A system might first identify the relevant concept and then select evidence according to source, temporal position, evidence type, or the requirements of the current task.

This does not require replacing conventional document or passage retrieval. Those mechanisms remain useful for identifying source material. Atomic organisation adds a complementary capability: the organisation of evidence through stable conceptual references. A software assistant responding to a question about an implementation decision, for example, could identify the relevant work-item concept and then retrieve the issue description, associated commits, and the developer discussion in which the design rationale was recorded.

The same structure is relevant to long-term memory in agentic systems. Memory stores based only on messages, summaries, observations, or embeddings can accumulate repeated and fragmented representations of the same underlying subject. An atomic memory architecture would instead attempt to associate new evidence with an existing persistent concept where appropriate and introduce a candidate concept when the current library does not adequately represent the new evidence.

The open-world principle is particularly important in this setting. An agent operating over time cannot assume that every future project, event, decision, or requirement belongs to a concept vocabulary fixed at deployment. A persistent memory system must therefore support both continuity and expansion: known concepts should accumulate relevant evidence, while genuinely new subjects should be representable without being forced into existing categories.

The present evaluation does not demonstrate that concept-organised retrieval improves question answering, reasoning accuracy, or retrieval quality. Nor does it measure semantic fidelity between evidence units and their associated concepts. These remain empirical questions. The current result establishes the underlying consolidation structure in the evaluated software corpora and therefore motivates, rather than validates, experiments on atom-centred retrieval and memory.

\subsection{LLMs as Dynamic Fusion Engines}
\label{sub:llmsasdynamic}

Within an atomic architecture, a large language model can be understood as a dynamic fusion engine. Its role need not be to serve simultaneously as the permanent repository of domain structure and the mechanism through which that structure is navigated. Instead, the model can interpret a task, identify relevant concepts, select supporting evidence, determine an appropriate composition, and express the resulting synthesis in a form adapted to the current context.

This view separates two functions that are often combined within a single generative process. The knowledge function concerns the maintenance of persistent concepts, relations, provenance, and reusable structures. The navigation function concerns selecting and combining those structures in response to a task. Existing architectural trends already exhibit partial forms of this separation. Chain-of-thought methods externalise intermediate reasoning states; tool-using systems invoke externally maintained operations rather than reconstructing all functionality internally \citep{schick2023toolformer,patil2023gorilla}; and modular or routed architectures select specialised computational components according to the current input \citep{sun2023modularity,pfeiffer2023modular}. Although these mechanisms differ, they share the broader principle that selection and navigation can be separated from the structures being selected.

The distinction can be stated directly in the notation of the Compression Calculus. Given a domain library
\[
A_D = \{a_1,a_2,\ldots,a_k\},
\]
the navigation problem is to identify an appropriate composition
\[
\alpha(\phi)=C(a_{i_1},a_{i_2},\ldots,a_{i_m})
\]
for the phenomenon or task $\phi$, subject to the available library and its composition rules. The representation problem is different: it concerns maintaining the library $A_D$ and its associated structure so that the resulting representations provide effective compression and reuse, expressed at domain level through $\mathrm{DCR}(D)$. The fusion-engine view therefore follows from the framework's existing distinction between the atomic library and the composition performed over that library.

A navigation analogy makes the separation intuitive. A navigation system is most useful when the road network exists independently of the mechanism that selects a route through it. Requiring the navigator to reconstruct the road network each time a destination is supplied conflates two problems of different kinds. The same distinction applies to knowledge architectures. Persistent domain structure can be maintained independently, while the model dynamically selects a path through that structure according to the current query.

Reasoning can then be interpreted, at least in part, as traversal and composition over a conceptual structure. In a clinical knowledge system, for example, a reasoning state might traverse from STEMI to a cath-lab protocol, anticoagulation, and post-PCI monitoring, while natural-language tokens provide the surface expression through which that traversal is requested and communicated. The important architectural point is not the particular domain sequence but the separation between a persistent conceptual structure and the dynamic process that selects a path through it.

This perspective also provides a structural interpretation of some generative failures. If navigation moves to a plausible but incorrect concept, or if the system combines units that should not compose under the domain's constraints, the generated continuation may remain linguistically plausible while becoming conceptually incorrect. The soft-atom analysis in Section~3.6 treats this as a problem of persistent identity, compositional stability, and compositional constraints rather than simply as a property of surface fluency.

In the evaluated software setting, the same distinction appears at a simpler level. A persistent work-item concept may be supported by an issue description, commits, comments, and developer discussions. Depending on the question, only part of this evidence may be relevant. A question concerning the origin of a defect may depend on early issue reports, while a question about the implementation may depend on commits and a question about design rationale may depend on developer discussion. The atomic layer preserves the common identity of the evidence, while the fusion engine performs task-specific selection.

Such an architecture also requires explicit controls. Evidence associated with one concept may conflict, become obsolete, or describe different stages of an evolving decision. The concept representation should therefore retain source, date, evidence type, temporal status, and relevant relations. A generated synthesis can then remain traceable to the evidence on which it depends.

Relations among atomic units are equally important. Concepts may depend on one another, supersede one another, conflict, or participate in broader processes. A mature atomic architecture should make these relations explicit rather than requiring the language model to infer the relevant structure independently on every invocation.

The resulting system is potentially more inspectable because the path through the conceptual structure can itself become part of the explanation. The system can expose which concepts were selected, which evidence supported them, and which relations were used to combine them. This does not by itself guarantee faithful reasoning, but it creates an explicit representational substrate against which navigation and synthesis can be evaluated.

The dynamic fusion-engine view therefore preserves the generative strengths of LLMs while assigning persistent knowledge organisation to a separate structure. LLMs remain central as interfaces, navigators, and synthesis mechanisms, while atomic libraries provide continuity, provenance, compositional constraints, and reusable conceptual identity.

\subsection{Self-Evolving Atomic Libraries}
\label{sub:selfevolving}

The open-world evaluation also motivates atomic libraries that can evolve as evidence accumulates. When a new source is introduced, some evidence will extend existing concepts, while other evidence may concern subjects not represented in the current library. A system that assumes a permanently closed vocabulary cannot distinguish these cases adequately.

The thread-level open-world model provides a simple operational example. An unlinked discussion is retained and permitted to introduce a candidate concept rather than being excluded from the corpus. This rule is deliberately coarse, but it demonstrates the broader principle that new evidence must remain capable of expanding the concept set.

A practical self-evolving library would require substantially more sophisticated concept-identity mechanisms. It would need to determine whether a candidate concept is genuinely new, whether it duplicates or overlaps an existing concept, whether multiple candidates should be merged, and whether an apparently unified concept should instead be divided. It would also need to revise representations when accumulated evidence shows that an earlier conceptual boundary was inadequate.

LLMs may assist this process by proposing concept descriptions, identifying recurring patterns, comparing new evidence with existing atomic units, and suggesting candidate relations. However, persistent concept creation should not depend solely on unconstrained generation. Changes to the atomic library affect subsequent retrieval and reasoning, so they should remain grounded in evidence and provenance and should be subject to consistency checks and, where required by the application, human review.

Usage information may provide an additional signal. Concepts that repeatedly organise relevant evidence may become stable components of the library. Concepts that consistently overlap may indicate unnecessary duplication. Concepts accumulating heterogeneous and weakly related evidence may require division. Recurrent combinations of lower-level concepts may also suggest candidates for higher-order atomic units.

This creates a form of cumulative knowledge development that differs from simply adding documents to a corpus or updating model parameters. New evidence changes not only the amount of stored information but potentially the structure through which that information is represented. The library itself therefore becomes an evolving object of learning.

The Compounding Cascade extends this principle beyond the first conceptual layer. Once stable atomic units have been identified, frequently recurring compositions may become candidate units at a higher level of abstraction. A self-evolving architecture would therefore need mechanisms not only for creating and revising concepts but also for identifying when repeated compositions justify new layers of representation.

The current evaluation provides only an initial foundation for this direction. It demonstrates that evidence can accumulate around persistent reference concepts and that the concept set must remain open when new sources are introduced. It does not yet provide an automatic atom-discovery algorithm or demonstrate that the resulting library grows according to any particular scaling law. These questions are addressed as testable predictions rather than as established findings.

\subsection{Scope and Future Work}
\label{sub:futurework}

The present evaluation measures evidence-to-concept consolidation across three software projects: HBase, Hadoop, and Maven. It does not implement the complete atom-centred architecture described in the theoretical sections. The measured concepts are operationally grounded in software work items and reply threads rather than being automatically discovered general-purpose atomic units. The evaluation also does not directly measure semantic fidelity, retrieval quality, automatic atom discovery, compositional gain, or the performance of higher-order abstraction layers.

Future work should first examine the semantic quality and practical usefulness of the consolidated concept structures. This includes determining whether a concept and its associated evidence support tasks such as software maintenance, issue understanding, change-impact analysis, technical question answering, and reconstruction of implementation decisions. Such evaluations should distinguish the correctness of the concept identity from the usefulness of the evidence organised around it.

A second direction concerns atomic retrieval. Experiments should compare conventional document- or chunk-based retrieval with retrieval organised through persistent concepts. Evaluation should measure not only whether relevant evidence is recovered, but also whether concept-organised retrieval improves completeness, reduces fragmentation, supports multi-hop questions, and preserves traceability under an equal evidence budget.

A third direction is automated concept discovery. The current open-world configuration uses reply threads as an operational estimate of concept identity for unlinked evidence. More advanced approaches could combine semantic similarity, temporal information, explicit trace links, repository metadata, and graph structure to identify candidate concepts. Such methods should preserve the open-world principle that evidence not explained adequately by the current library remains capable of expanding the concept set.

A fourth direction concerns concept description and consolidation. Persistent identifiers connect evidence but do not by themselves provide complete representations of concept meaning. Future work should investigate how concise concept descriptions can be generated, updated, validated, and linked to the evidence from which they are derived, while preserving distinctions among source, temporal state, and provenance.

A fifth direction is temporal evolution. Software concepts change as requirements are revised, problems are redefined, and implementations are replaced. An atomic library should therefore distinguish between additional evidence concerning a stable concept and evidence indicating that the concept itself has changed. Versioned atomic representations and temporal relations may be necessary for maintaining accurate histories.

A sixth direction concerns explicit compositional structure. The present evaluation measures consolidation but not CCG or higher-order layer compression. Future studies should examine how persistent concepts combine into larger structures and whether the layer-quality effects formalised in Section~4.8 can be estimated empirically. In particular, no $q_i$ is estimated for the current corpora, and the measured consolidation ratios should not be interpreted as approximations of an unknown optimum.

The framework is also being applied to enterprise knowledge bases at SeKondBrain AI Lab through a feature-graph extraction pipeline. Dedicated feature-composition and extraction analyses are being developed to measure consolidation at successive levels of the abstraction structure. These measurements will be reported separately once the evidence-unit definitions, concept-identity rules, and associated evaluation procedure can be specified in sufficient detail for the resulting quantities to be interpreted and reproduced.

Finally, the evaluation should be extended to additional software corpora and to domains with different recurrence structures and different forms of persistent identity. The contrasting results considered in Section~5.5 already indicate that consolidation should not be expected to be universal. Different domains will require different evidence units and concept models, but the methodological requirements remain the same: unit boundaries must be explicit, the concept space must be capable of expansion, and every representation must remain connected to its underlying evidence.

\subsection{Testable Predictions}
\label{sub:predictions}

The framework leads to several predictions that can be tested independently of the present consolidation measurements. None is treated as validated by the current evaluation.

\begin{enumerate}

\item A system with an explicit atomic library will outperform a substantially larger model without such a library on domain-specific tasks when task definition and access to external evidence are held constant. One concrete test would compare an atom-centred system with a baseline model approximately an order of magnitude larger while providing both systems with the same evidence corpus and evaluation tasks. The prediction is falsified if the larger model matches or exceeds the atom-centred system under those controlled conditions.

\item Achievable compression is bounded by the information content of the represented domain. Under the uniquely decodable encoding assumptions of Equation~(4.15), a measured atomic representation should not exceed the bound $\mathbb{E}|s(X)|/H(X)$. This prediction can be tested in domains for which both the surface encoding and domain entropy can be estimated. A reproducible measured ratio above the corresponding theoretical bound would falsify either the prediction or the assumptions used to instantiate the encoding.

\item Improvements in atomic quality at earlier abstraction layers will have multiplicative rather than purely additive effects on end-to-end representational performance. Equation~(4.36) predicts that changes in the layer-quality factors $q_i$ propagate through the cascade. An experiment could vary the quality of one early atomic layer while holding later procedures fixed and compare its effect across cascades of different depth. The prediction is falsified if the effect of an equivalent early-layer improvement remains invariant as cascade depth increases.

\item The size of a useful self-evolving atomic library will grow sub-linearly relative to the accumulation of surface evidence once stable recurring structures have been discovered. A stronger form of the prediction is that library growth approaches logarithmic behaviour while evidence continues to grow approximately linearly. The released software corpora provide a direct test: mailing-list evidence can be introduced month by month while recording $|C|$ and the number of retained evidence units. Sustained linear growth of the concept library under a stable identity procedure would falsify the stronger prediction.

\item Retrieval organised around persistent concepts will outperform strong chunk-based retrieval on questions requiring evidence distributed across several artefacts. A suitable benchmark would construct multi-hop questions whose supporting evidence is distributed across issues, commits, comments, and developer discussions, while holding the total retrieved evidence budget constant. The prediction is falsified if a strong chunk-retrieval baseline matches concept-organised retrieval on answer quality, evidence completeness, and traceability.

\item Expert conceptual libraries will exhibit comparable orders of magnitude across mature professional domains. The chess literature reviewed in Section~2.1 motivates a provisional scale of approximately $50{,}000$ learned units, but the framework does not assume this quantity to be universal. The prediction can be tested by operationally defining reusable units in domains such as legal practice and financial trading and estimating the stable expert library under consistent criteria. The prediction is falsified if comparable expert domains exhibit no clustering around a common order of magnitude.

\end{enumerate}

These predictions separate claims that follow from the framework from results already established by the current experiments. The present study measures consolidation under explicit concept and evidence definitions. Subsequent work can therefore test whether the same representational principles also produce improvements in retrieval, compositional reasoning, library growth, and system-level performance.

\section{Conclusion}
\label{sec:conclusion}

This paper has argued that scalable knowledge representation depends not only on the amount of information available to a system, but also on the level at which that information is organised. Documents, messages, files, and other surface artefacts preserve evidence, provenance, and context, but information concerning the same underlying subject is often distributed across many such artefacts. Atomic units provide an intermediate representational layer through which distributed evidence can be organised around persistent and reusable conceptual structures.

The central theoretical contribution is the Compression Calculus, which formalises the relationship between surface evidence and atomic representation. The framework distinguishes observed evidence units, $U$, from persistent reference concepts, $C$, and operationalises evidence-to-concept consolidation through

\[
R = \frac{|U|}{|C|}.
\]

It further develops measures of atomic and domain compression, compositional compression, layered abstraction, and atomic quality, together with the Compounding Cascade thesis, according to which representational effects can propagate multiplicatively across successive abstraction layers.

The empirical evaluation applies the consolidation component of this framework to three software projects using repository evidence from SEOSS-33 and public Apache mailing-list archives. Under the thread-level open-world concept model and at deduplicated paragraph granularity, HBase contains $664,480$ evidence units associated with $29,026$ concepts, producing a consolidation ratio of $22.89{:}1$ and a reduction of $95.6\%$. Hadoop contains $949,131$ evidence units associated with $47,187$ concepts, producing a ratio of $20.11{:}1$ and a reduction of $95.0\%$. Maven, by contrast, contains $98,004$ paragraph-level evidence units associated with $8,571$ concepts, producing a ratio of $11.43{:}1$ and a reduction of $91.3\%$. The $20{:}1$ target is therefore reached in the two larger projects at paragraph granularity, but not in the smaller Maven corpus and not at message granularity in any of the three projects.

These results establish a boundary on the empirical claim. Consolidation is not a universal constant of software evidence and should not be interpreted independently of corpus scale, evidence-unit definition, recurrence structure, or concept-identity rules. The analysis further shows that recurrence introduced specifically by the second source is not the principal driver of the reported reductions. Instead, much of the effect is already present within the underlying repository evidence and is amplified by the paragraph-level unit definition. The multi-source contribution is therefore methodological: heterogeneous evidence from repositories and developer communication can be retained and organised under a common open-world concept model without restricting the corpus to material that already fits a predefined identifier system.

The contrasting corpora considered in the boundary analysis reinforce this qualification. MeetingBank exhibits only $1.06{:}1$ consolidation under its agenda-item concept model, demonstrating that substantial evidence-to-concept recurrence is not present in every domain or corpus. MedMentions likewise produces considerably lower consolidation than the software corpora under curator-assigned UMLS concept identity. These comparisons indicate that consolidation depends on the representational task and the type of concept being modelled rather than reflecting a domain-independent compression law.

The open-world treatment of concept identity remains an important methodological contribution. Evidence that cannot be linked to a pre-existing JIRA identifier is retained rather than discarded and may introduce additional thread-level concepts. Both $U$ and $C$ can therefore change as new evidence is incorporated. In the evaluated software setting, this produces a more conservative ratio than the corresponding closed-world configuration because previously unlinked evidence is allowed to expand the denominator rather than being excluded.

The evaluation should nevertheless be distinguished from the stronger theoretical notion of a fully developed atomic unit. JIRA issue identifiers provide persistent identity, while reply-thread concepts provide a weaker operational form of continuity. The current experiments do not establish compositional stability or explicit compositional constraints for these concepts. They therefore measure consolidation around persistent reference structures rather than demonstrating that every operational concept satisfies all properties of an atomic unit.

This distinction connects the empirical findings to the paper's analysis of the meaning gap in contemporary generative systems. Transformer representations can encode substantial syntactic and semantic structure, but the paper characterises context-dependent conceptual approximations that lack persistent identity, stable composition, or explicit compositional constraints as soft atoms. The argument is not that large language models lack useful internal representations, but that persistent conceptual identity constitutes a different architectural property from reconstructing task-relevant structure dynamically within a context.

This motivates the separation between representation and navigation developed in the discussion. Within an atom-centred architecture, persistent conceptual structures provide continuity, relations, provenance, and reusable domain knowledge, while an LLM operates as a dynamic fusion engine that interprets a task, selects relevant atomic units and supporting evidence, determines an appropriate composition, and expresses the resulting synthesis. The model therefore acts as a navigator over an evolving conceptual structure rather than being required to serve simultaneously as both the structure and the mechanism that traverses it.

The open-world results further motivate self-evolving atomic libraries. As new evidence arrives, some observations should extend existing concepts while others should introduce new ones. A mature system must therefore support concept creation, revision, merging, separation, and higher-order composition while preserving links to the original evidence. The current thread-based procedure provides only an operational approximation of this process, but it demonstrates why a closed vocabulary is insufficient when knowledge sources evolve.

Several claims of the broader framework remain untested. The present evaluation measures consolidation but does not directly measure semantic fidelity, retrieval quality, automatic atom discovery, compositional compression gain, or end-to-end improvements in reasoning. The cross-domain cascades developed in the paper are analytical estimates under stated assumptions rather than empirical measurements. Similarly, the predictions concerning atomic-library efficiency, entropy-bounded compression, multiplicative quality effects, sub-linear library growth, concept-organised retrieval, and expert library scale remain hypotheses for future evaluation.

Future work should therefore move from measuring the existence of consolidation structure to testing its usefulness. This includes comparing concept-organised retrieval with strong chunk-based baselines, developing methods for discovering and validating candidate atomic units, modelling temporal evolution and relations between concepts, and evaluating whether persistent conceptual structures improve multi-hop reasoning, semantic memory, technical question answering, decision reconstruction, and retrieval-augmented generation. Further evaluation across additional software projects and other domains will also be necessary to determine how consolidation varies with corpus scale, recurrence structure, evidence granularity, and concept definition.

In conclusion, the paper contributes a formal account of atomic representation, an open-world methodology for measuring evidence-to-concept consolidation, and an architectural interpretation of how persistent conceptual structures may interact with generative models. The empirical results show that consolidation above $20{:}1$ can occur at scale in two large software corpora under an explicitly defined paragraph-level, open-world concept model, while Maven and the contrasting MeetingBank results demonstrate that the effect is conditional rather than universal. The broader contribution is therefore not a claim that one compression ratio characterises intelligence, but that the choice, persistence, quality, and composition of representational units provide a measurable and testable layer between raw evidence and intelligent processing. Developing systems that can discover, maintain, navigate, and progressively refine such units is the next step towards practical atomic knowledge architectures.

\section*{Acknowledgment}
The authors gratefully acknowledge Benjamin Brey (SeKondBrain), Priyanka Kochhar (SeKondbrain) \& Dr Sharon Jheeta (outside advisor) for reading early versions of this paper and providing insightful editorial suggestions.

\section*{AI Assistance \& Provenance}
The author directed the research question, developed the central thesis, made all substantive methodological decisions, and accepts full responsibility for the manuscript, results, claims, citations, and errors.
AI tools were used as follows: 
\begin{itemize}
    \item Manus (December 2025): research synthesis, literature checking, outline refinement, editing, consistency review, and reference validation. 
    \item Claude by Anthropic (February-March 2026): conceptual exploration, research assistance, source discovery, structural suggestions, and integration of reviewer feedback. 
    \item ChatGPT by OpenAI (March 2026): phrasing and abstract variants, and literature scoping to identify candidate sources. 
    \item OpenAI Codex (July-August 2026): assistance implementing and reviewing experimental scripts, processing public datasets and checking numerical consistency.
\end{itemize}
The author reviewed the AI processed material, independently checked cited sources, reran the reported experiments, and approved the final manuscript. No confidential, customer, or personally identifiable data were intentionally provided to these tools.
\begin{Backmatter}







\bibliography{bibo}

\end{Backmatter}

\end{document}